\documentclass[a4paper]{cas-dc}

\usepackage[numbers,sort&compress]{natbib}
\usepackage{hyperref}
\hypersetup{colorlinks=false}
\usepackage{amsthm}
\usepackage{bm}
\usepackage{algorithm}
\usepackage{algorithmic}

\usepackage{adjustbox, pifont, booktabs, multirow, placeins, float, indentfirst, tabularx, amsmath, amssymb}

\newdefinition{definition}{Definition}

\newcommand{\method}{Diff-PC}
\newcommand{\TrueTitle}{Diff-PC: Identity-preserving and 3D-aware Controllable Diffusion for Zero-shot Portrait Customization}

\begin{document}
\let\printorcid\relax
\def\floatpagepagefraction{1}
\def\textpagefraction{.001}

\shorttitle{\TrueTitle}

\shortauthors{Y. Xu et al.}

\title[mode=title]{\TrueTitle}

\author[1]{Yifang Xu}
\ead{xyf@smail.nju.edu.cn}
\credit{Conceptualization, Data curation, Formal analysis, Investigation, Methodology, Software, Visualization, Writing -- original draft}

\author[1]{Benxiang Zhai}
\credit{Data curation, Validation, Writing -- review \& editing}

\author[2]{Chenyu Zhang}
\credit{Data curation, Investigation, Methodology, Visualization}

\author[3]{Ming Li}
\credit{Investigation, Methodology, Writing -- review \& editing}

\author[1]{Yang Li}
\credit{Conceptualization, Investigation, Funding acquisition, Supervision}

\author[1]{Sidan Du}
\ead{coff128@nju.edu.cn}
\credit{Funding acquisition, Project administration, Supervision, Writing -- review \& editing}

\affiliation[1]{
    organization={
        School of Electronic Science and Engineering, Nanjing University
    },
    city={Nanjing},
    postcode={210036},
    country={China}
}

\affiliation[2]{
    organization={
        School of Business, Hohai University
    },
    city={Nanjing},
    postcode={211100},
    country={China}
}

\affiliation[3]{
    organization={
        School of Artificial Intelligence, Nanjing University of Information Science and Technology
    },
    city={Nanjing},
    postcode={210044},
    country={China}
}

\cormark[1]

\cortext[1]{Corresponding author}

\begin{abstract}
Portrait customization (PC) has recently garnered significant attention due to its potential applications. However, existing PC methods lack precise identity (ID) preservation and face control. To address these tissues, we propose \textbf{\method}, a \textbf{diff}usion-based framework for zero-shot \textbf{PC}, which generates realistic portraits with high ID fidelity, specified facial attributes, and diverse backgrounds. Specifically, our approach employs the 3D face predictor to reconstruct the 3D-aware facial priors encompassing the reference ID, target expressions, and poses. To capture fine-grained face details, we design ID-Encoder that fuses local and global facial features. Subsequently, we devise ID-Ctrl using the 3D face to guide the alignment of ID features. We further introduce ID-Injector to enhance ID fidelity and facial controllability. Finally, training on our collected ID-centric dataset improves face similarity and text-to-image (T2I) alignment. Extensive experiments demonstrate that \method\space surpasses state-of-the-art methods in ID preservation, facial control, and T2I consistency. Furthermore, our method is compatible with multi-style foundation models.
\end{abstract}




\maketitle

\section{Introduction}
\label{sec:intro}

Text-to-image (T2I) \cite{LDM-2021-SD, Imagen-2022, DiT-2022, SDXL-2023, SD3-2024, ControlNet-2023} generation technology has witnessed remarkable advancements in the past few years, thanks to the rapid development of large-scale text-image datasets \cite{LAION-5B-2022, DiffusionDB-2023} and diffusion models \cite{DDPM-2020, DDIM-2020, Improved-Diffusion-2021, LDM-2021-SD, Imagen-2022, DiT-2022, SDXL-2023, SD3-2024} (e.g., DDPM \cite{DDPM-2020}, Stable Diffusion \cite{LDM-2021-SD}, DiT \cite{DiT-2022}). As a subtask of T2I, portrait customization (PC) \cite{DreamBooth-2023, HyperDreamBooth-2023, IP-Adapter-2023, PhotoMaker-2023, FlashFace-2024, PuLID-2024, ConsistentID-2024, HiFi-Portrait-2025} garnered considerable attention due to its immense potential in applications like virtual try-on \cite{TryOnDiffusion-2023, GP-VTON-2023}, e-commerce advertising \cite{sivathanuCustomersOnlineShopping2023, gaoArtificialIntelligenceAdvertising2023}, and portrait animation \cite{ID-Animator-2024, LivePortrait-2024}. Given a reference portrait and additional prompts, PC aims to generate realistic human images retaining the reference identity (ID) while other attributes adhere to the prompts. 

Early PC works \cite{Pirenderer-2021, SofGAN-2022, 3DFaceShop-2023} typically employ GANs \cite{GAN-2014}, which can be viewed as a combination of face reenactment \cite{HyperReenact-2023} and editing \cite{DiffusionRig-2023}. However, these methods are limited in both image diversity and ID fidelity. To meet the user demand for high-quality PC, numerous diffusion-based methods \cite{DreamBooth-2023, HyperDreamBooth-2023, IP-Adapter-2023, PhotoMaker-2023, FlashFace-2024, PuLID-2024, ConsistentID-2024} have emerged, enhancing the diversity and accuracy of personalized portraits. DreamBooth \cite{DreamBooth-2023} and HyperDreamBooth \cite{HyperDreamBooth-2023} propose fine-tuning the UNet \cite{U-Net} or LoRA \cite{LoRA-2021} during inference to learn the corresponding ID information. Although achieving decent results, fine-tuning for each ID takes from tens of minutes to several hours, and the synthesized face similarity depends on the quality of manually collected data. These factors significantly hinder the practical application of such methods.

\begin{figure}[b!]
  \centering
  \includegraphics[width=\linewidth]{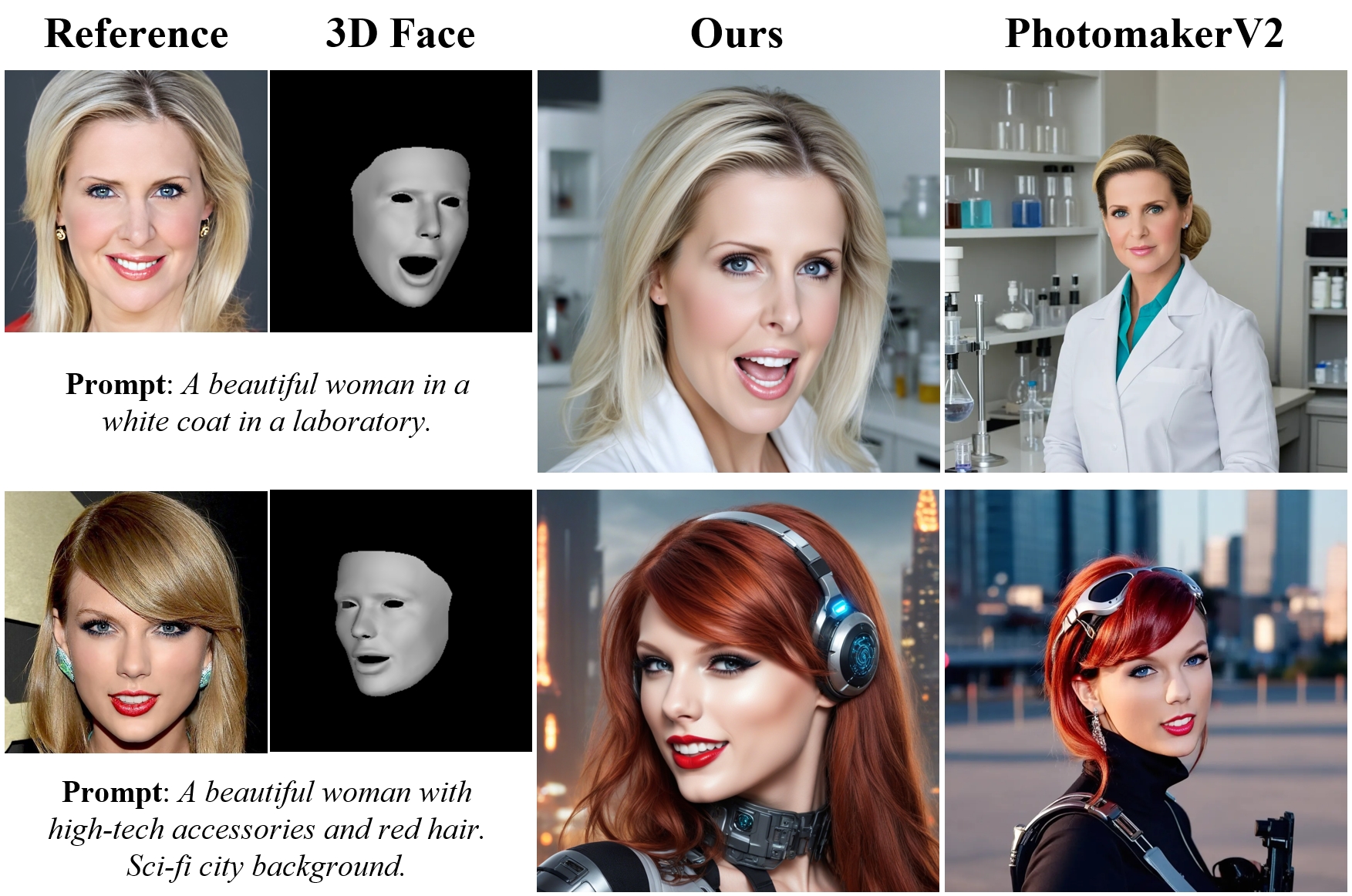}
  \caption{
    \textbf{Two examples of portrait customization.} Compared to the SOTA method (PhotoMaker-V2), our approach achieves superior identity fidelity, text-to-image consistency, and the ability to specify facial position and expression.
  }
  \label{fig:example}
\end{figure}

\begin{figure*}[t!]
  \centering
  \includegraphics[width=\linewidth]{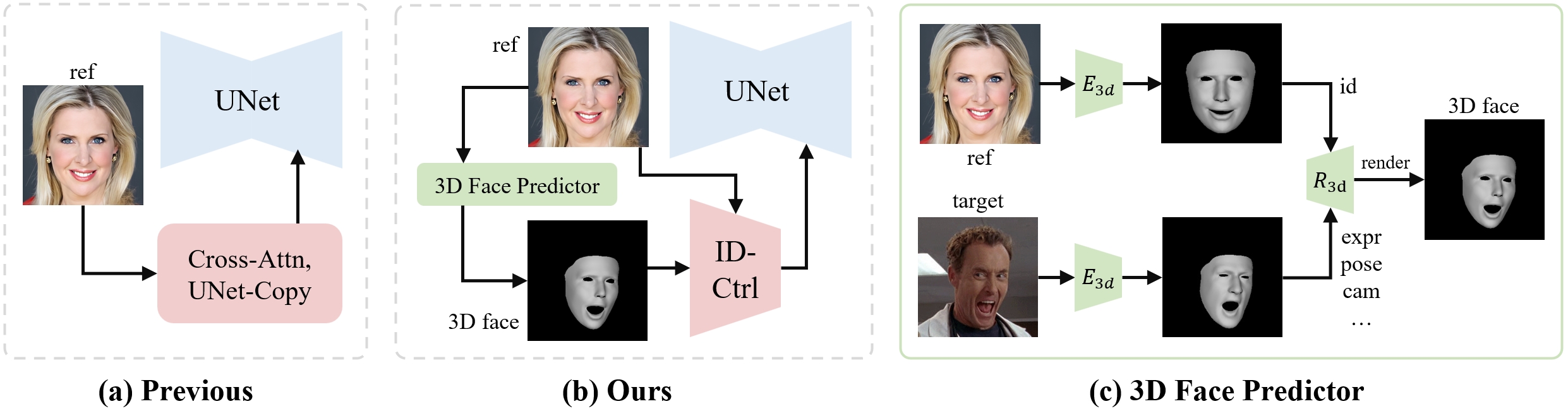}
  \caption{
    \textbf{(a) Previous methods} lack facial geometric conditions, obtaining poor results. \textbf{(b) Our approach} leverages the 3D-aware facial prior to preserve identity details and control facial attributes. \textbf{(c) 3D face predictor} synthesizes the 3D face that integrates the reference identity with the target expression and pose.
  }
  \label{fig:high-level}
\end{figure*}

To obviate the need for fine-tuning during inference, recent methodologies \cite{IP-Adapter-2023, PhotoMaker-2023, FlashFace-2024, PuLID-2024, ConsistentID-2024, HiFi-Portrait-2025} introduce a zero-shot (tuning-free) setup capable of generating images within seconds. Specifically, IP-Adapter \cite{IP-Adapter-2023, IP-Adapter-FaceID-2024} and PhotoMakerV2 \cite{PhotoMaker-2023, PhotoMaker-V2-2024} pre-train cross-attention \cite{Transformer} mechanisms to enable the model to comprehend face semantics. FlashFace \cite{FlashFace-2024} replicates and pre-trains an additional UNet to inject the reference ID, as illustrated in Fig.~\ref{fig:high-level}~(a). Nevertheless, these approaches lack facial control conditions, rendering them incapable of precisely customizing facial expressions and poses, as shown in the rightmost column of Fig.~\ref{fig:example}. Furthermore, they neglect fine-grained facial features, thereby compromising ID fidelity. Additionally, the generated images often exhibit inconsistencies with text prompts.

To overcome the aforementioned challenges, we propose a tuning-free framework for portrait customization, named \method, as depicted in Fig.~\ref{fig:high-level}~(b). In contrast to existing methods, our approach accurately customizes portrait attributes (such as facial expression, position, accessory, and background) while achieving high ID fidelity. Some examples can be found in Fig.~\ref{fig:example}. Concretely, we first employ the 3D face predictor to reconstruct the 3D face that integrates the reference ID with the target expression and pose, as depicted in Fig.~\ref{fig:high-level}~(c). To obtain more refined ID features, we design the ID-Encoder that fuses global and local facial embeddings. Subsequently, we devise the identity-preserving controller (ID-Ctrl), using 3D-aware face priors as the guiding condition, which embeds and aligns ID features to the latent space. In addition, we introduce the ID-Injector to enhance face similarity and controllability. Moreover, we develop an automated pipeline to collect a high-quality ID-centric dataset. Training our model with this dataset assists in reinforcing face similarity and mitigating T2I misalignment. Finally, our model is compatible with many different styles of SDXL foundation models. In conclusion, our primary contributions are as follows:

\begin{itemize}
\item 
We propose \textbf{Diff-PC}, a diffusion-based framework for zero-shot portrait customization, which generates ID-preserved portraits with diverse facial expressions and poses in different backgrounds.
\item 
We design the ID-Ctrl module, leveraging a 3D-aware facial prior to guide the alignment of ID features, thereby achieving high ID fidelity while controlling facial attributes.
\item 
We collect an ID-centric dataset to specifically train our model, enhancing face similarity and mitigating T2I inconsistency.
\item 
Extensive qualitative and quantitative experiments demonstrate that our method attains state-of-the-art performance in ID fidelity, facial controllability, and T2I alignment.
\end{itemize}

\section{Related Work}
\label{sec:related_work}

\subsection{Text-to-image Generation}
\label{re:t2i}
Generating realistic images from a given textual description, initially proposed by AlignDRAW \cite{T2I-1st-work-AlignDRAW-2016}, presents a challenging task. Early T2I researches \cite{GAN-T2I-2016, StackGAN-2017, DM-GAN-2019, wang2024scgan} primarily employ generative adversarial networks (GANs) \cite{GAN-2014}, which are limited to creating images within specific close-set domains \cite{dataset-Caltech-UCSD-birds-2010, dataset-Microsoft-COCO-2014} and limited textual conditions, seriously hindering their applicability. 

With the advancement of diffusion models \cite{DDPM-2020, DDIM-2020, Improved-Diffusion-2021, zhuDiffusionbasedDiverseAudio2024, xu2023query} and large-scale text-image datasets \cite{LAION-5B-2022, DiffusionDB-2023}, there has been a surge of diffusion-based T2I methods \cite{LDM-2021-SD, SDXL-2023, SD3-2024, DiT-2022, Imagen-2022, Custom-Diffusion-2023}, markedly improving the quality and diversity of T2I generation. A representative open-source work is Stable Diffusion (SD) \cite{LDM-2021-SD}, which uses CLIP \cite{CLIP-2021} to extract text embeddings and feeds it into cross-attention mechanisms to guide the denoising process. Furthermore, the entire process is conducted within the latent space encoded by VAE, significantly reducing computational cost. ControlNet \cite{ControlNet-2023} trains a replicated UNet \cite{U-Net} encoder to control SD to support additional input conditions, including depth maps, body keypoints, facial landmarks, etc. Based on SD, SDXL \cite{SDXL-2023} enhances generation performance by extending the U-Net \cite{U-Net} architecture and introduces an extra text encoder \cite{CLIP-2021} to improve text-image consistency. This paper builds upon SDXL and is compatible with multiple style models based on SDXL.

\subsection{3D Face Reconstruction}
Compared to 2D portraits, 3D face models \cite{zhaoFacialExpressionTransfer2024} offer more details, including ID shape, facial expressions, wrinkles, albedo, lighting, etc. In the past two decades, numerous works \cite{BFM-2009, FaceWarehouse-2013, FLAME-2017, LSFM-2018, DILP-2024, yuPedestrian3dShape2024} have emerged in the field of monocular 3D face reconstruction, focusing on predicting the parameters of established 3D morphable models (3DMM) \cite{3DMM-1999}, e.g. BFM \cite{BFM-2009} and FLAME \cite{FLAME-2017}. Some approaches optimize 3D modeling through analysis-by-synthesis, yet re-optimizing parameters for each new image incurs substantial computational costs. To overcome this issue, recent 3DMM-based studies \cite{DECA-2021, EMOCA-2022, HiFace-2023, DreamFace-2023, TokenFace-2023, SMIRK-2024}, such as DECA \cite{DECA-2021} and SMIRK \cite{SMIRK-2024}, leverage deep learning for efficient 3D reconstruction. In addition, these techniques can decouple various 3D parameters, such as ID, expression, and pose.

This paper employs SMIRK \cite{SMIRK-2024} to selectively extract and replace parameters related to expression and pose for predicting geometry meshes. This process enables face attribute transfer without requiring albedo and lighting. Subsequently, we render the 3D geometry and project it onto 2D images as guiding conditions for diffusion model \cite{DDIM-2020}. Similar works include DiffusionAct \cite{DiffusionAct-2024}, DiffSwap \cite{DiffSwap-2023}, and RichDreamer \cite{qiu2024richdreamer}, which leverage 3D-aware conditions (e.g., 3D face, keypoints and landmarks) to align with face features to guide the denoising process, achieving face reenactment \cite{DiffusionAct-2024}, face swapping \cite{DiffSwap-2023}, and depth estimation \cite{PFANet-2021, DAFFNet-2021}, respectively. Specifically, DiffusionAct employs landmarks in face reenactment, which helps improve cross-viewpoint consistency, ensuring that facial features remain consistent across different viewpoints. DiffSwap uses 3D faces for face swapping to enhance ID fidelity. However, unlike these approaches, this paper focuses on tuning-free portrait customization with high ID fidelity, face controllability, and T2I consistency.

\subsection{Portrait Customization}
Given a reference image and a textual prompt, portrait customization (PC) aims to synthesize T2I human images that retain the reference ID. Early GAN-based PC approaches \cite{Pirenderer-2021, SofGAN-2022, 3DFaceShop-2023} can be considered a combination of face reenactment and editing, but its generation suffers from low quality and diversity. Moreover, these methods lose the high-frequency details from reference, compromising ID preservation in the synthesized faces. Subsequent tuning-based researches \cite{Textual-Inversion-2022, DreamBooth-2023, HyperDreamBooth-2023, FaceChain-2023, EasyPhoto-2023, Moment-GPT-2025} addresses these issues using diffusion models, which operate in latent space and better maintain the high-frequency features from the reference face. Textual Inversion \cite{Textual-Inversion-2022} trains text embeddings to preserve ID, but yields faces with low resemblance to the original. DreamBooth \cite{DreamBooth-2023} fine-tunes UNet to associate the facial ID with specific rare words, but involves a large number of trainable parameters. FaceChain \cite{FaceChain-2023} and EasyPhoto \cite{EasyPhoto-2023} mitigate this problem by training LoRA \cite{LoRA-2021}, which has fewer parameters. However, these tuning-based methods require individual fine-tuning during inference for each person, consuming tens of minutes to several hours, which significantly hinders their practical application. Furthermore, the ID fidelity of the generated portraits is greatly influenced by the quality of training images.

\subsection{Zero-shot Portrait Customization}
To eliminate the need for fine-tuning during inference, recent approaches \cite{IP-Adapter-2023, PhotoMaker-2023, FlashFace-2024, PuLID-2024, ConsistentID-2024, PhotoVerse-2D-2023, FaceSnap-2025, Face2Diffusion-2024, VTG-GPT-2023, MM-Diff-2024, IDAdapter-2024, liLearningAdversarialSemantic2024, HP3-2025, HiFi-Portrait-2025} present a zero-shot (tuning-free) paradigm that generates portraits in mere seconds. These methods typically employ large-scale face recognition datasets for pre-training, such as CelebA \cite{dataset-CelebA-2015}, FFHQ \cite{FFHQ-2019}, and LAION-Face \cite{FaRL-2022}. IP-Adapter \cite{IP-Adapter-2023} pre-trains cross-attention layers to enable the model to comprehend facial semantics. Building upon IP-Adapter, PhotoMakerV2 \cite{PhotoMaker-2023, PhotoMaker-V2-2024} fuses face embeddings with class embeddings (e.g., man and woman) to enhance T2I alignment. FlashFace \cite{FlashFace-2024} duplicates and pre-trains an additional UNet to inject reference ID. PuLID \cite{PuLID-2024} introduces the SDXL-Lightning \cite{SDXL-Lightning-2024} branch and contrastive learning to improve ID preservation.

Despite the success of these tuning-free methods in preserving ID, they are unable to precisely control expressions and poses via text prompts. Moreover, due to training on noisy facial datasets and the absence of detailed facial features, ID fidelity is compromised. To address above challenges, this paper proposes \method, which leverages 3D-aware face priors in ID-Ctrl to guide and align fine-gained facial features, thereby achieving precise face control and superior ID retention. Additionally, we collect an ID-centric dataset to further enhance facial similarity and T2I consistency.

\section{Preliminary}
\label{sec:Preliminary}
Stable Diffusion (SD) \cite{LDM-2021-SD} is a widely adopted open-source generative model consisting of VAE \cite{VAE-2014}, time-conditional denoising UNet $\epsilon_{\theta}$, and CLIP \cite{CLIP-2021} text encoder $E_{txt}$. By operating within a compressed latent space rather than the pixel space, SD enables efficient training and inference. More specifically, given a target image $I_{tgt} \in \mathbb{R}^{H \times W \times 3}$, SD initially employs the VAE encoder $E_{vae}$ to compress $I_{tgt}$ into a latent code $z_{0} = E_{vae}(I_{tgt}) \in \mathbb{R}^{\frac{H}{8} \times \frac{W}{8} \times 4}$. Subsequently, the diffusion process adds noise to $z_{0}$ to generate the noisy latent $z_{t}$ with
\begin{equation}
\label{eq:z_t}
    z_t = \sqrt{\bar{\alpha}_t} z_0 + \sqrt{1 - \bar{\alpha}_t} \epsilon
\end{equation}
\begin{equation}
    \bar{\alpha}_{t} = \prod_{s=1}^{t} \alpha_{s}
\end{equation}
where $t \sim \text{Uniform}(\{1, . . . , T \})$ denotes the timestep at which noise is introduced. $\epsilon \sim \mathcal{N}(0,1)$ is the ground-truth noise sampled from a standard Gaussian distribution. $\{\alpha_s\}$ represents a set of pre-defined scaling factors that control the amount of noise added at each timestep $s$. Finally, SD learns $\epsilon_{\theta}$ to estimate the noise added at the $t$-th timestep. The overall training objective is defined as:
\begin{equation}
    \mathcal{L} = \mathbb{E}_{z_t, t, c_{txt}, \epsilon} [|| \epsilon - \epsilon_{\theta}(z_t, t, c_{txt})||_{2}]
\label{eq:diff_loss}
\end{equation}
where text embedding $c_{txt} = E_{txt}(P_{txt})$ is obtained by encoding text prompt $P_{txt}$ through $E_{txt}$. SD leverages cross-attention mechanisms to capture the semantics of $c_{txt}$.

\begin{figure*}[t!]
  \centering
  \includegraphics[width=0.95\linewidth]{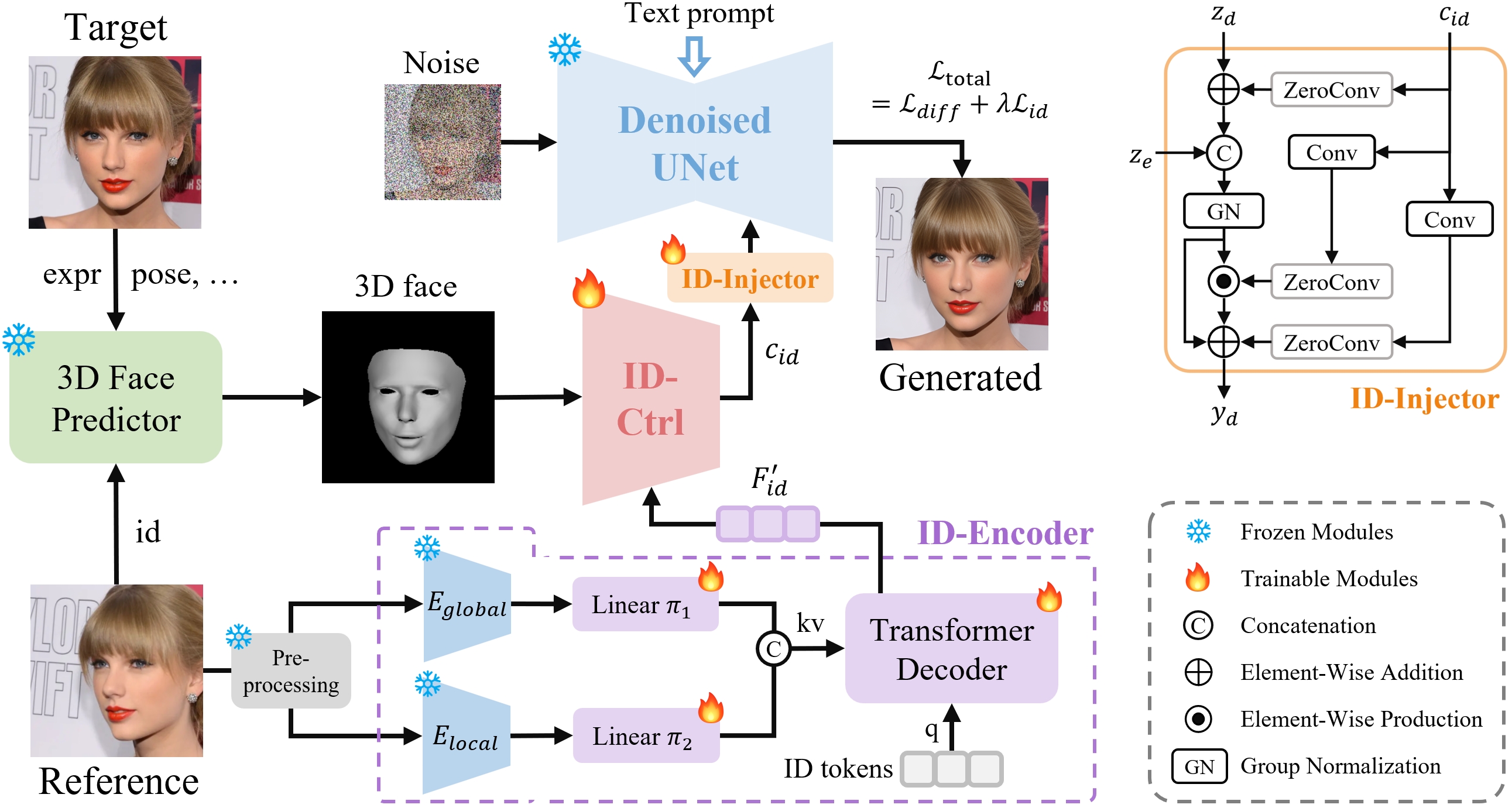}
  \caption{
    \textbf{The overall training architecture of \method}, which is built upon SDXL (VAE is omitted in the figure). First, we employ the 3D face predictor to reconstruct a 3D facial prior $I_{3d}$ that contains reference ID and target attributes (e.g., expressions and postures). Next, fine-grained ID features $F_{id}^{\prime}$ are obtained via ID-Encoder. Subsequently, we utilize $I_{3d}$ to guide the alignment of $F_{id}^{\prime}$ in ID-Ctrl. Finally, we inject $c_{id}$ into UNet through ID-Injector to accomplish ID preservation and facial control.
  }
  \label{fig:training}
\end{figure*}

\section{Method}
\label{sec:method}
In Sec.~\ref{subsec:overview}, we present an overview of our model. Sec.~\ref{subsec:3d_face} and \ref{subsec:id_encoder} explain 3D face predictor and ID-Encoder, respectively. In Sec.~\ref{subsec:id_ctrl}, we detail our key components: ID-Ctrl (identity-preserving controller) and ID-Injector. Sec.~\ref{subsec:train_infer} covers training and inference. Finally, Sec.~\ref{subsec:dataset} introduces the ID-centric dataset we collected.

\subsection{Overview}
\label{subsec:overview}

Given a reference image $I_{ref} \in \mathbb{R}^{H \times W \times 3}$, \method\ aims to synthesize a realistic human image $I_{gen} \in \mathbb{R}^{H \times W \times 3}$ that preserves the identity (ID) of $I_{ref}$. The generated face's pose and expression consistent with the target image $I_{tgt} \in \mathbb{R}^{H \times W \times 3}$, while other attributes, such as accessories and background, are specified by the text prompt $P_{txt}$. The above process can be simplified as: 
\begin{equation} 
    I_{gen} = M(I_{ref}, I_{tgt}, P_{txt}) 
\end{equation} 
where $M$ denotes our \method. Its overall training architecture is based on the pre-trained SDXL \cite{SDXL-2023}, as shown in Fig.~\ref{fig:training}. For convenience of explanation, VAE is omitted from the figure. To be specific, we reconstruct the 3D-rendered face $I_{3d} \in \mathbb{R}^{H \times W}$ using the 3D face predictor. And extract high-fidelity ID features $F_{id}^{\prime} \in \mathbb{R}^{32 \times d}$ by ID-Encoder (with pre-processing). Subsequently, we design ID-Ctrl $M_{ctrl}$ to obtain ID condition $c_{id}$, which contains reference ID and facial control information. Finally, we employ ID-Injector to infuse $c_{id}$ into UNet denoiser $\epsilon_{\theta}$, improving ID fidelity and facial controllability.

\subsection{3D Face Predictor}
\label{subsec:3d_face}
Fig.~\ref{fig:high-level}~(c) depicts the framework of our 3D face predictor. We first feed $I_{ref}$ and $I_{tgt}$ into the FLAME-based \cite{FLAME-2017} SMIRK \cite{SMIRK-2024} encoder $E_{3d}$ to regress the facial matrix $H \in \mathbb{R}^{5023 \times 3}$, which consist of identity $\alpha$, expression $\beta$, and pose $\gamma$ parameters. Among these, $\beta$ adjusts facial expressions, eye closure, and jaw rotation, while $\gamma$ controls the orthographic camera coefficients and rigid pose. This pipeline is formulated as: 
\begin{gather}
    (\alpha_{ref}, \beta_{ref}, \gamma_{ref}) = E_{3d}(I_{ref}) \\
    (\alpha_{tgt}, \beta_{tgt}, \gamma_{tgt}) = E_{3d}(I_{tgt})
\end{gather}
Next, utilizing the differentiable rasterizer $R_{3d}$ in PyTorch3D \cite{Pytorch3D-2020}, we combine the reference identity $\alpha_{ref}$, target expression $\beta_{tgt}$ and pose $\gamma_{tgt}$ to generate the monochrome rendering of the reconstructed 3D face $I_{3d} \in \mathbb{R}^{H \times W}$ with
\begin{equation}
    I_{3d} = R_{3d}(\alpha_{ref}, \beta_{tgt}, \gamma_{tgt})
\end{equation}

\subsection{Pre-processing and ID-Encoder}
\label{subsec:id_encoder}

\paragraph{Pre-processing.}
For the reference image $I_{ref}$, we leverage BackgroundMattingV2 \cite{BackgroundMatting-2020} for human matting to eliminate cluttered backgrounds. Subsequently, RetinaFace \cite{RetinaFace-2020} $M_{det}$ is utilized for face detection to get the largest facial region $I_{face} \in \mathbb{R}^{H_1 \times W_1 \times 3}$.

\paragraph{ID-Encoder.}
After obtaining $I_{face}$, we design ID-Encoder to obtain ID features $F_{id}^{\prime} \in \mathbb{R}^{32 \times d}$, as depicted in Fig.~\ref{fig:training}. We first employ pre-trained networks \cite{ArcFace-2019, CLIP-2021} to derive multi-granular and complementary facial features from $I_{face}$: global facial embedding $F_{global} \in \mathbb{R}^{1 \times d_1}$ and local facial embedding $F_{local} \in \mathbb{R}^{257 \times d_2}$. $F_{global}$ is encoded via ArcFace \cite{ArcFace-2019} $E_{global}$, containing critical ID information (e.g., face shape and skin tone) that facilitates distinguishing the different individuals. In contrast, $F_{local}$ captures intricate details (such as skin texture) and is extracted by the image encoder of CLIP \cite{CLIP-2021} $E_{local}$. Then, we project $F_{global}$ and $F_{local}$ into a common dimension $d$ using linear layers $\pi_{1}$ and $\pi_{2}$, respectively.
\begin{gather}
    F_{global}^{\prime} = \pi_{1}[E_{global}(I_{face})] \\
    F_{local}^{\prime} = \pi_{2}[E_{local}(I_{face})]
\end{gather}
Subsequently, we apply a vanilla transformer \cite{Transformer} decoder $M_{dec}$ to fuse $F_{global}^{\prime}$ and $F_{local}^{\prime}$, producing refined ID features $F_{id}^{\prime}$. This fusion captures fine-grained ID details, enhancing the ID fidelity of synthesized images. The process is formalized as:
\begin{gather}
    q = F_{id},\quad kv = F_{global}^{\prime} \mathbin{\textcircled{c}} F_{local}^{\prime} \\
    F_{id}^{\prime} = M_{dec}(q, kv)
\end{gather}
Here, $F_{id} \in \mathbb{R}^{32 \times d}$ is learnable ID tokens serving as \textit{query} $q$ of $M_{dec}$. $kv$ denotes \textit{key} and \textit{value} of $M_{dec}$. $\mathbin{\textcircled{c}}$ represents the concatenation operation.

\begin{figure}[t!]
  \centering
  \includegraphics[width=\linewidth]{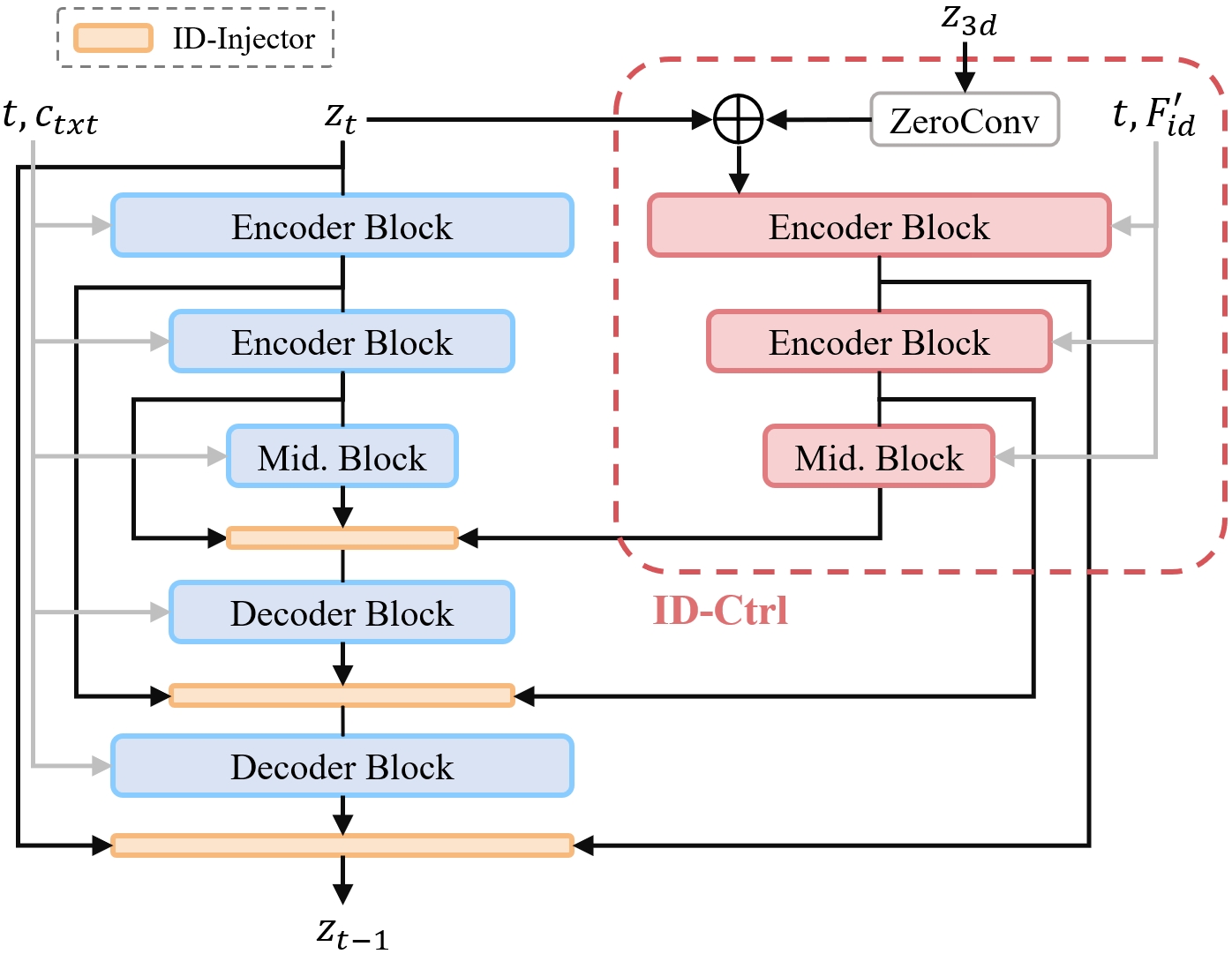}
  \caption{
    \textbf{ID-Ctrl and ID-Injector in \method.} For brevity, we only draw two blocks in the encoder and decoder, but they actually have three blocks each.
  }
  \label{fig:id-ctrl}
\end{figure}

\subsection{ID-Ctrl and ID-Injector}
\label{subsec:id_ctrl}
\paragraph{ID-Ctrl.} 
After getting $I_{3d}$ and $F_{id}^{\prime}$, inspired by ControlNet \cite{ControlNet-2023}, we use $I_{3d}$ as image-space conditioning input and encode $I_{3d}$ into the latent code $z_{3d}$ via $E_{vae}$. Subsequently, we develop ID-Ctrl (see Fig.~\ref{fig:id-ctrl}), whose trainable backbone is replicated from the UNet encoder and middle blocks. For simplicity in illustration, we only draw two blocks for the encoder and decoder, but in fact they both have three blocks. We then utilize $z_{3d}$ as the input condition, which guides $F_{id}^{\prime}$ to align with $z_{3d}$ at the latent space. This process is implemented through the cross-attention \cite{Transformer} in ID-Ctrl:
\begin{gather}
    Q=\Omega_{q}(z_{3d}),\quad   K=\Omega_{k}(F_{id}^{\prime}),\quad V=\Omega_{v}(F_{id}^{\prime}) \\
    \text{Attention}(Q, K, V) = \text{softmax} \left(\frac{QK^{T}}{\sqrt{d_k}}\right) V
\end{gather}
where $Q$, $K$, and $V$ denote the \textit{query}, \textit{key}, and \textit{value}, respectively. $\Omega_{q}$, $\Omega_{k}$, and $\Omega_{v}$ are projection layers. Finally, ID-Ctrl $M_{ctrl}$ outputs $c_{id}$ with
\begin{equation}
    c_{id} = M_{ctrl}(z_{t}, t, z_{3d}, F_{id}^{\prime})
\end{equation}
$c_{id}$ encapsulates ID details and facial control information. Next, it is injected into UNet via ID-Injector.

\paragraph{ID-Injector.}
The original ControlNet \cite{ControlNet-2023} utilizes zero convolutions to add control conditions into the UNet directly. However, this approach is overly simplistic, as shown below:
\begin{equation}
    z_{d}^{\prime} = \text{GN}[z_{e} \mathbin{\textcircled{c}} (z_{d} + c_{id})]
\end{equation}
Here, $z_{d}$ represents the output of the middle or decoder blocks in UNet, and $z_{e}$ is the skip connection from encoder blocks. GN denotes group normalization \cite{GroupNorm-2018}. Inspired by SFT \cite{SFT-2018}, we devise ID-Injector to introduce affine transformation parameters $\phi$ and $\mu$, enhancing facial control in the generated images. The architecture of ID-Injector is presented in the upper right corner of Fig.~\ref{fig:training}, and it can be formulated as:
\begin{gather}
    \phi = \text{Conv}_{1}(c_{id}),\quad  \mu = \text{Conv}_{2}(c_{id}) \\
    y_{d} = (1 + \phi) \textcircled{\footnotesize$\cdot$} z_{d}^{\prime} + \mu
\end{gather}
where $y_{d}$ is the output of ID-Injector. \textcircled{\footnotesize$\cdot$} denotes element-wise multiplication. Fig.~\ref{fig:id-ctrl} illustrates the placement of ID-Injector within UNet; it is positioned after the middle and decoder blocks.

\subsection{Training and Inference}
\label{subsec:train_infer}

\paragraph{Training.}
Based on the ID-centric dataset we constructed, during the training phase, we randomly select two images from the same ID as the reference and target image. The optimization objective of \method, denoted as $\mathcal{L}_{diff}$, is similar to that of Stable Diffusion (Eq.~\ref{eq:diff_loss}). But we additionally introduce $c_{id}$ as a control condition, formulated as:
\begin{equation}
    \mathcal{L}_{diff} = || \epsilon - \epsilon_{\theta}(z_t, t, c_{txt}, c_{id})||_{2}
\end{equation}
Moreover, we employ the ID loss $\mathcal{L}_{id}$ to improve ID fidelity \cite{DiffSwap-2023, IDAdapter-2024}. According to Eq.~\ref{eq:z_t}, we directly recover $\hat{z}_{0}$ (the predicted initial code) from $z_{t}$ and $\epsilon_{\theta}$ using:
\begin{equation}
    \hat{z}_0 = \frac{z_t - \sqrt{1 - \bar{\alpha}_t} \epsilon_{\theta}}{\sqrt{\bar{\alpha}_t}}
\end{equation}
Next, we apply VAE decoder $D_{vae}$ to obtain $I_{gen} = D_{vae}(\hat{z}_0)$. The ID loss $\mathcal{L}_{id}$ is defined as:
\begin{equation}
\label{eq:id_loss}
    \mathcal{L}_{id} = M_{det}(I_{gen}) \cdot (1 - \text{cos}[E_{global}(I_{ref}), E_{global}(I_{gen})])
\end{equation}
Here, $M_{det}$ is RetinaFace \cite{RetinaFace-2020} used for face detection; it outputs 1 if a face is detected and 0 otherwise. The inputs $I_{ref}$ and $I_{gen}$ to $E_{global}$ are only the largest facial regions. Finally, the total loss (with $\lambda = 0.1$) is given by:
\begin{equation}
\label{eq:total_loss}
    \mathcal{L}_{total} = \mathcal{L}_{diff} + \lambda \mathcal{L}_{id}
\end{equation}

\begin{figure}[t!]
  \centering
  \includegraphics[width=\linewidth]{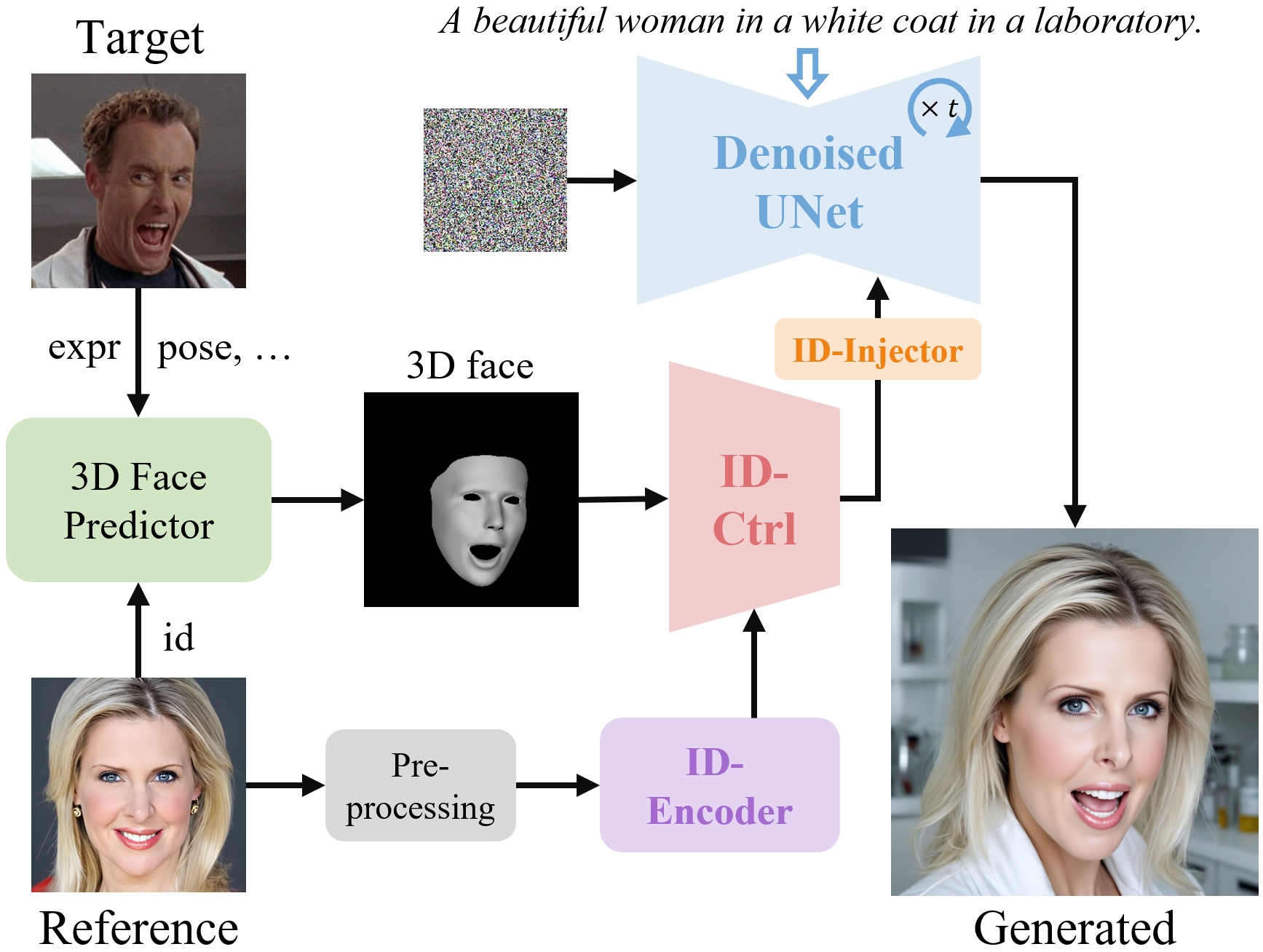}
  \caption{
    The inference pipeline.
  }
  \label{fig:inference}
\end{figure}

\paragraph{Inference.}
During the inference stage, we employ $I_{ref}$ and $I_{tgt}$ from different individuals, as illustrated in Fig.~\ref{fig:high-level}~(c) and \ref{fig:inference}. The denoising process at each timestep is expressed by the following equation:
\begin{equation}
    z_{t-1} = \frac{1}{\sqrt{\alpha_t}} \left( z_t - \frac{1 - \alpha_t}{\sqrt{1 - \bar{\alpha}_t}} \epsilon_\theta\right) + \sigma_t \epsilon
\end{equation}
where $\sigma_t$ denotes the preset noise scaling factor. After $t$ iterations, we obtain the final generated image $I_{gen} = D_{vae}(z_0)$.

\subsection{ID-centric Dataset}
\label{subsec:dataset}
In our initial experiments, we train \method\space on the popular face dataset (LAION-Face \cite{FaRL-2022}), where the reference and target are the same images. However, we obtain sub-optimal results due to:  (1) excessive noisy images in LAION-Face. (2) The inconsistency of target images between training and inference. A straightforward solution is to use open-source datasets \cite{dataset-VGGFace2-2018, dataset-IMDb-Face-2018, dataset-WebFace260M-2021} that contain multiple images per ID. However, these datasets are typically designed for face recognition tasks, where the face is aligned and occupies a relatively fixed proportion of the image.

To address these issues, we develop a pipeline to create an ID-centric dataset containing high-quality and diverse text-portrait pairs. Training our \method\space on this dataset enhances ID fidelity, facial controllability, and text-to-image (T2I) consistency. Moreover, we believe this dataset will facilitate future research in portrait customization.

\paragraph{Data gathering.}
Our dataset is based on IMDB-Face \cite{dataset-IMDb-Face-2018} that provides URLs to the non-aligned original images, from which we downloaded all valid entries. To improve the diversity of identities in the dataset, we curate a list of names from DouBan\footnote{https://movie.douban.com}, whose names do not appear in IMDB-Face. According to this list, we gather about 100 images per person on the Internet.

\paragraph{Data processing.}
We first filter out images where the number of faces is not equal to 1, which is achieved by detecting facial bounding boxes via RetinaFace \cite{RetinaFace-2020}. Next, we crop each image by randomly enlarging these bounding boxes, ensuring that the face area occupies at least 2\% of the cropped image. Additionally, the cropped images with sides smaller than $256$ are removed, and the rest are resized to $1024 \times 1024$. To enhance dataset quality, we utilize the FQA model\footnote{https://www.modelscope.cn/models/iic/cv\_manual\_face-quality-assessment\_fqa} to evaluate facial quality, discarding images with scores lower than 0.5. PaddleOCR\footnote{https://github.com/PaddlePaddle/PaddleOCR} is employed to identify watermarks on faces, and any images containing watermarks are deleted.

\paragraph{ID verification.}
To determine if an image matches a specific ID, we start by using ArcFace \cite{ArcFace-2019} to obtain facial embeddings. Subsequently, the k-means clustering algorithm is applied to select the central embedding from the largest cluster as the representative embedding. Ultimately, only images with a similarity greater than 0.7 to the representative embedding are retained, effectively filtering out images from other IDs.

\paragraph{Image captioning.}
To ensure consistency between text prompts and generated images, we acquire detailed image descriptions using Yi-Vision \cite{Yi-2024}. Specifically, we utilize the following instruction: "\textit{Dear Yi-Vision, please provide a detailed description of this portrait, focusing on the subject's appearance, hairstyle, clothing, accessories, and image background.}"

\paragraph{Data statistics.}
Finally, applying the above process, we construct a training dataset containing 32k IDs and 650k images, averaging 20 images per ID, as shown in Tab.~\ref{tab:dataset_image_id}. Additional dataset statistics can be found in App.~\ref{app:dataset}.

\begin{table}[t!]
\centering
\caption{
    Statistics of identities and images in our dataset. It originates from two parts: IMDB-Face \cite{dataset-IMDb-Face-2018} and Internet (data we collect on the web).
}
\label{tab:dataset_image_id}
\begin{adjustbox}{max width=\linewidth}
\begin{tabular}{@{}cccc@{}}
\toprule
\textbf{Dataset} & IDs & Imgs & Imgs/ID \\ 
\midrule
IMDB-Face & 21k & 390k & 19 \\
Internet & 11k & 250k & 23 \\
All & 32k & 650k & 20 \\ 
\bottomrule
\end{tabular}
\end{adjustbox}
\end{table}

\section{Experiments}
\label{sec:experiments}

\subsection{Implement Details}
Our method is built on SDXL-Base-1.0 \cite{SDXL-2023, SDXL-based-1-2024}, using training images with a $1024 \times 1024$ resolution. The version of ArcFace \cite{ArcFace-2019} and CLIP \cite{CLIP-2021} image encoder is antelopev2 and OpenCLIP ViT-H/14, respectively. During training, we only update the linear layers and transformer decoder in ID-Encoder, as well as ID-Ctrl and ID-injector, while keeping the parameters of other components frozen. To improve generation quality, there is a 5\% probability of dropping the text prompt. Moreover, there is a 5\% probability of setting the global or local facial embedding to the all-zero matrix. We use the Adam \cite{AdamW-2017} optimizer with a fixed learning rate 1e-5. The model is trained over 700k iterations with a batch size 20, utilizing 4 NVIDIA A800 GPUs. In the inference phase, we employ SDXL-Lightning \cite{SDXL-Lightning-2024} and DDIM \cite{DDIM-2020} sampler with 8 steps.

\begin{figure*}[t!]
  \centering
  \includegraphics[width=\linewidth]{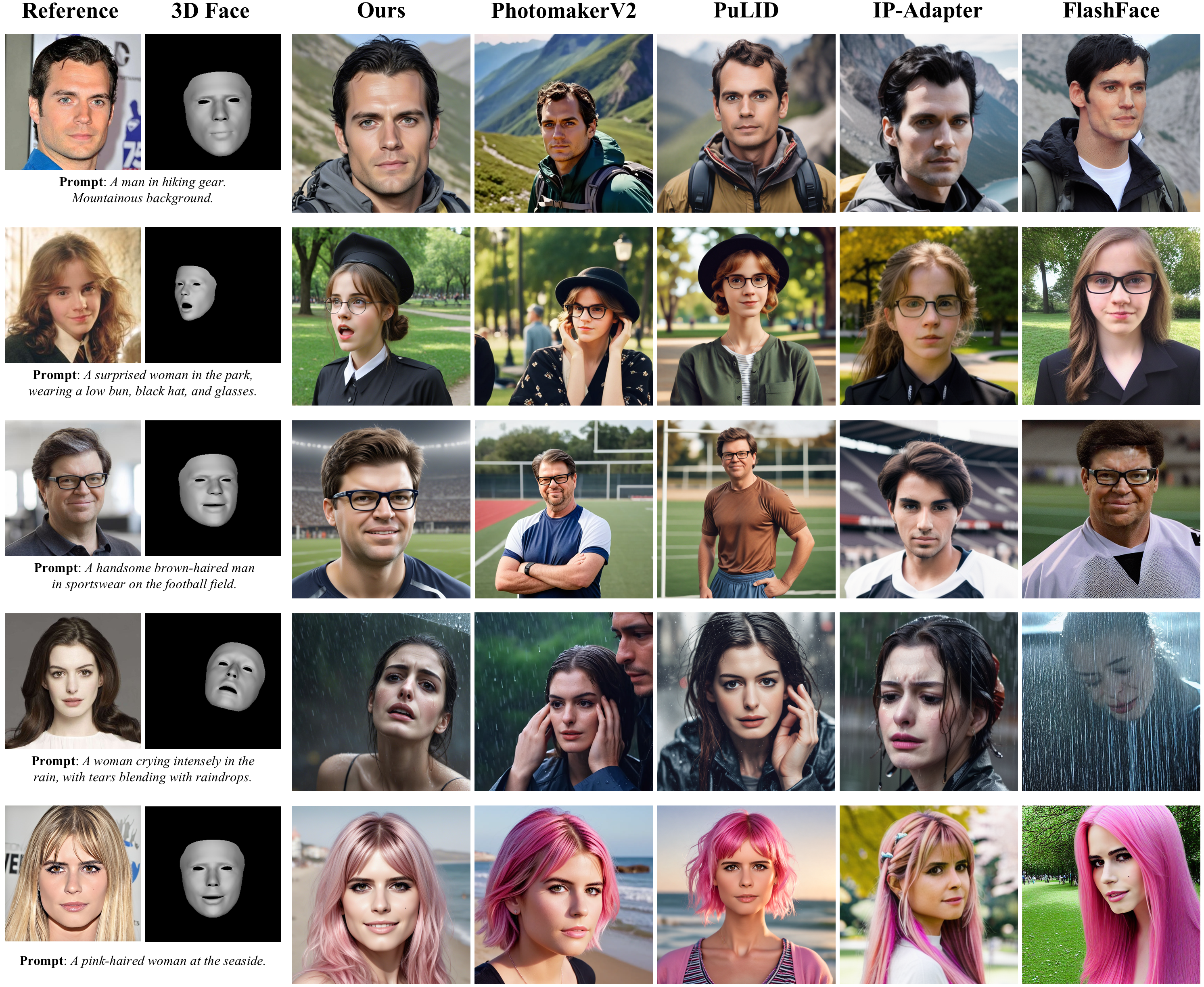}
  \caption{
    \textbf{Qualitative results.} Compared to the SOTA methods, our framework achieves high-quality portrait customization, offering superior ID fidelity while maintaining aesthetic appeal. Additionally, it allows precise control through specified 3D-aware face and text prompt. 
  }
  \label{fig:compare}
\end{figure*}

\begin{figure*}[t!]
  \centering
  \includegraphics[width=\linewidth]{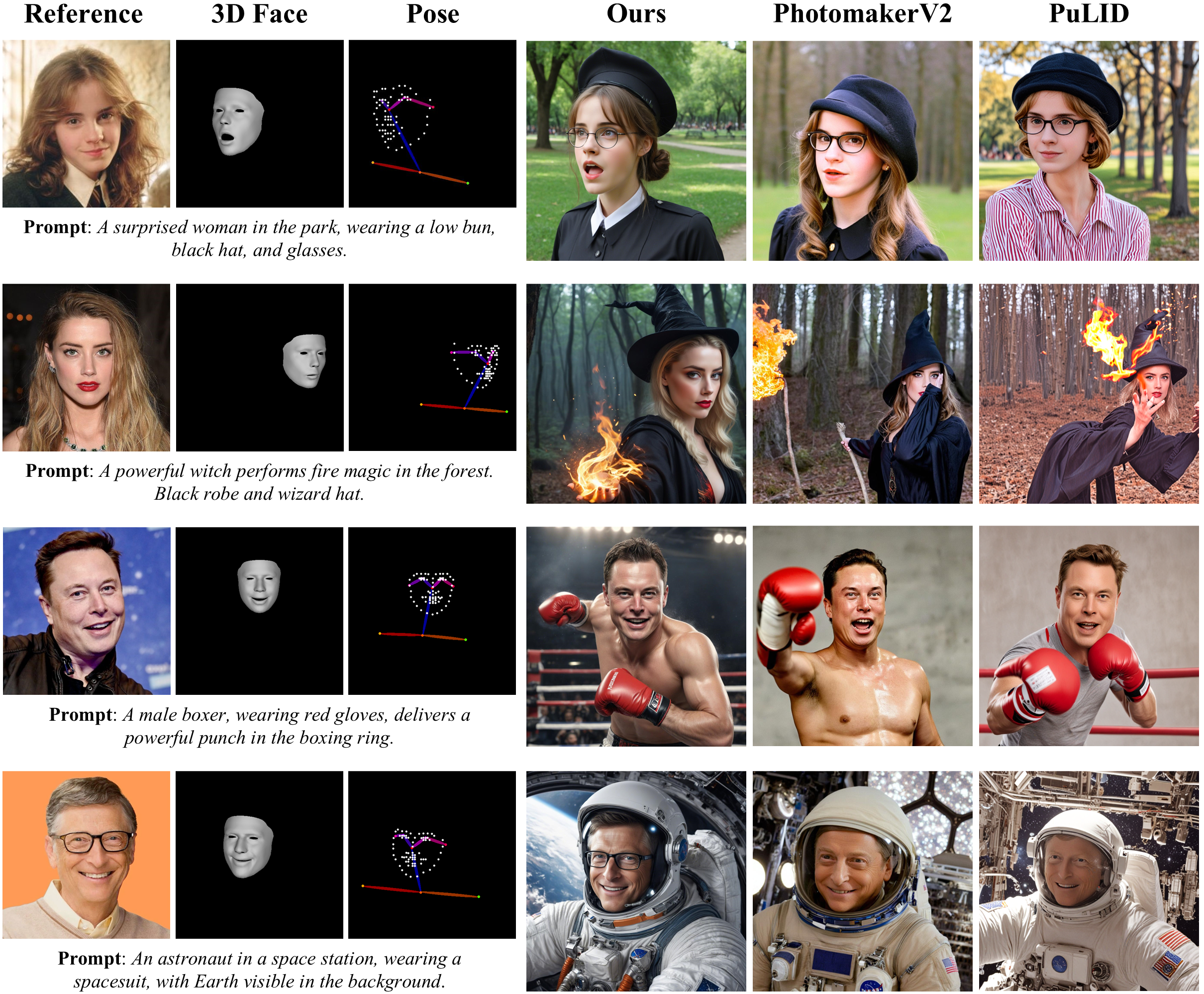}
  \caption{
    \textbf{Qualitative comparison} of \method\space with other approaches integrating Pose-ControlNet.
  }
  \label{fig:compare_ctrl}
\end{figure*}

\begin{figure*}[t!]
  \centering
  \includegraphics[width=\linewidth]{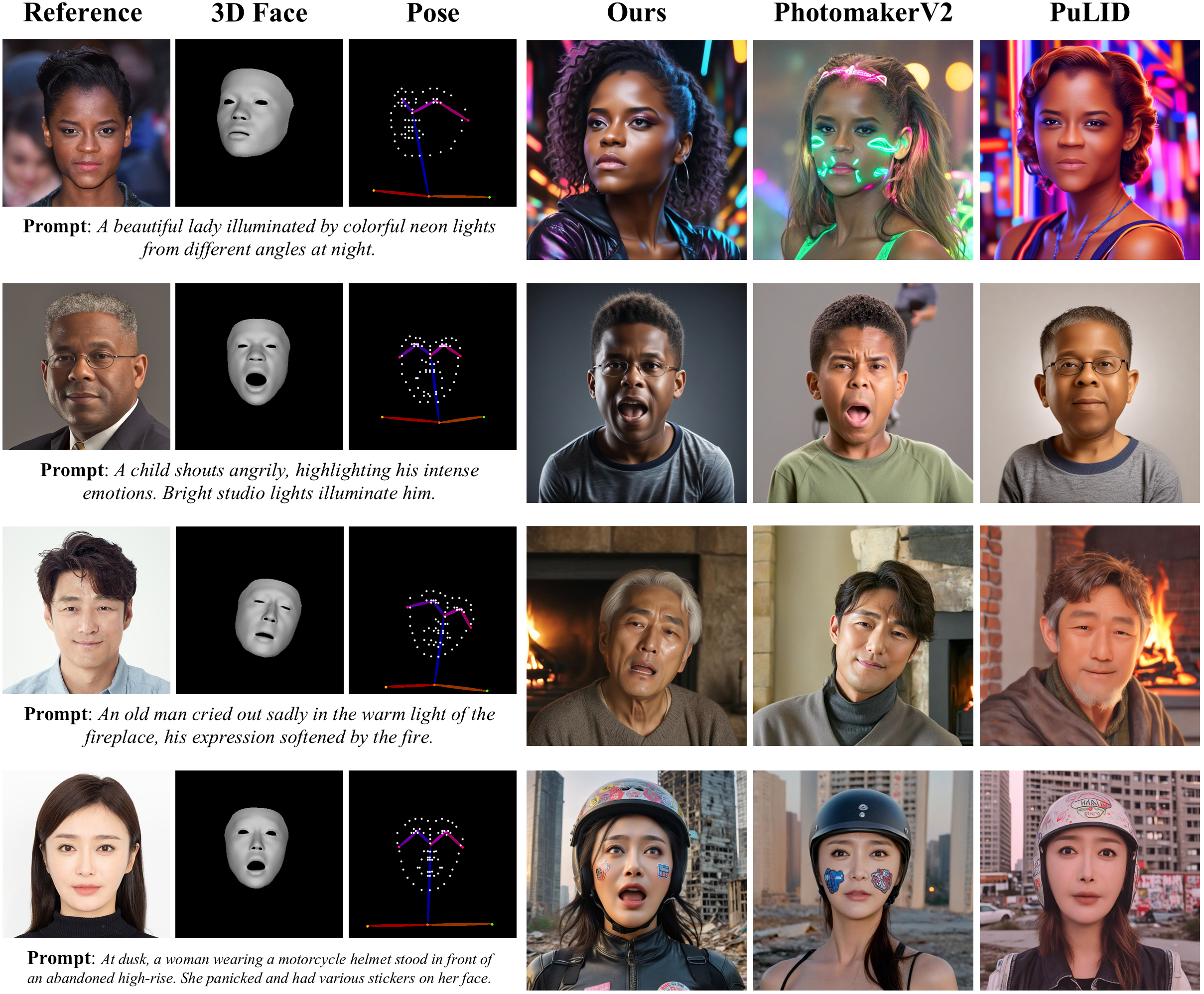}
  \caption{
    \textbf{Qualitative comparison}.
  }
  \label{fig:compare_ctrl_2}
\end{figure*}

\subsection{Qualitative Analysis}
\paragraph{Qualitative comparison}
This paper primarily focuses on zero-shot portrait customization (PC). In Fig.~\ref{fig:compare}, we compare our approach with several open-source state-of-the-art (SOTA) methods, including FlashFace \cite{FlashFace-2024}, IP-Adapter \cite{IP-Adapter-2023}, PuLID \cite{PuLID-2024}, and PhotoMaker \cite{PhotoMaker-2023}. To ensure fairness, we use the best-performing versions of IP-Adapter-FaceID \cite{IP-Adapter-FaceID-2024} and PhotoMakerV2 \cite{PhotoMaker-V2-2024}. As evident from the figure, our method outperforms others in real-world scenes, achieving superior ID preservation, aesthetics, and precise control over facial attributes (such as pose, position, and expressions). In contrast, other techniques fail to maintain ID fidelity, visual appeal, and face manipulation. Comparing rows 2 to 4, our generated images adhere more closely to the given text prompts, whereas other methods show text-to-image (T2I) inconsistencies. Overall, given one reference image, \method\space generates realistic portraits with strong ID retention while offering customizable and diverse face attributes, accessories, and backgrounds.

\paragraph{Face controlability}
For a more fair comparison, we integrate Pose-ControlNet \cite{ControlNet-2023, ControlNet-v1_1-2023} with target conditions (including face landmarks and body poses) into previous methods to achieve facial control, as shown in Fig.~\ref{fig:compare_ctrl} and \ref{fig:compare_ctrl_2}. The results reveal that previous methods, even with Pose-ControlNet, struggle to accurately control facial expressions and poses. This issue is primarily due to the lack of detailed face geometry information in pose images. In contrast, our method leverages the 3D-aware face to guide the denoising process, enhancing facial control. Additionally, comparing the 1st row of Fig.~\ref{fig:compare_ctrl} with the 2nd row of Fig.~\ref{fig:compare} demonstrates that using Pose-ControlNet in previous methods reduces ID fidelity. \method\space addresses these challenges through the designed ID-Ctrl and ID-Injector modules, highlighting its facial control capabilities.

\begin{table*}[t!]
\caption{
    \textbf{Quantitative comparison} of Diff-PC against the SOTA methods on ID-centric-200. All metrics are reported as percentages (\%). The best-performing results are indicated in \textbf{bold}.
}
\label{tab:qualitative}
\centering
\begin{adjustbox}{max width=0.8\linewidth}
\begin{tabular}{@{}ccccccccc@{}}
\toprule
\multirow{2}{*}{\textbf{Methods}} & \multicolumn{3}{c}{\scriptsize ID fidelity} & \multicolumn{2}{c}{\scriptsize Face control} & \scriptsize T2I & \multicolumn{2}{c}{\scriptsize Others} \\ 
\cmidrule(lr){2-4} \cmidrule(lr){5-6} \cmidrule(lr){7-7} \cmidrule(lr){8-9}
& $Sim \uparrow$ & $CLIP_{i}\uparrow$ & $Shape\downarrow$ & $Expr\downarrow$ & $Pose\downarrow$ & $CLIP_{t}\uparrow$ & $FID\downarrow$ & $Time\downarrow$ \\ \midrule
FlashFace \cite{FlashFace-2024} & 68.3 & 70.6 & 16.0 & 21.0 & 7.8 & 28.4 & 338.1 & 5.4s \\
IP-Adapter \cite{IP-Adapter-FaceID-2024} & 69.1 & 67.5 & 15.8 & 19.6 & 7.1 & 29.8 & 341.0 & \textbf{4.4s} \\
PuLID \cite{PuLID-2024} & 70.9 & 66.8 & 14.7 & 20.5 & 7.7 & 30.3 & 328.3 & 4.9s \\
PhotoMaker-V2 \cite{PhotoMaker-V2-2024} & 71.2 & 69.7 & 15.1 & 19.3 & 6.9 & 30.5 & 325.1 & 4.6s \\
\textbf{Diff-PC} (Ours) & \textbf{74.1} & \textbf{72.4} & \textbf{12.3} & \textbf{16.9} & \textbf{5.6} & \textbf{31.1} & \textbf{314.4} & 5.7s \\ \bottomrule
\end{tabular}
\end{adjustbox}
\end{table*}

\subsection{Quantitative Analysis}
\paragraph{Evaluation datasets}
To evaluate the quantitative results, we construct three benchmark datasets. Specifically, we randomly select 200 IDs from our ID-centric dataset that are not used during training, referred to as ID-centric-200. From the FFHQ \cite{FFHQ-2019} dataset, we randomly chose 150 IDs, denoted as FFHQ-150. Additionally, we utilize the Unsplash-50 \cite{Unsplash-50-2024} dataset, which comprises 50 IDs uploaded to the Unsplash website between February and March 2024.

For each ID, we create an ID combination consisting of one reference image, three target images, three text prompts, and three random seeds. The target images are randomly chosen from LAION-Face \cite{FaRL-2022}, and prompts are generated by GPT-4 \cite{GPT-4}. Notably, the target images and prompts for each ID combination are entirely different. For instance, for ID-centric-200, we synthesize a total of $5400 = 200 \times 3 \times 3 \times 3$ different images.

\begin{table*}[t!]
\caption{
    \textbf{Quantitative results} on FFHQ-150 and Unsplash-50.
}
\label{tab:ffhq_unsplash}
\centering
\begin{adjustbox}{max width=\linewidth}
\begin{tabular}{@{}ccccccccccccccc@{}}
\toprule
\multirow{3}{*}{\textbf{Methods}} & \multicolumn{7}{c}{\textbf{FFHQ-150}} & \multicolumn{7}{c}{\textbf{Unsplash-50}} \\ 
\cmidrule(l){2-8} \cmidrule(l){9-15}
 & \multicolumn{3}{c}{\scriptsize ID fidelity} & \multicolumn{2}{c}{\scriptsize Face control} & \scriptsize T2I & \scriptsize Quality & \multicolumn{3}{c}{\scriptsize ID fidelity} & \multicolumn{2}{c}{\scriptsize Face control} & \scriptsize T2I & \scriptsize Quality \\ 
\cmidrule(l){2-4} \cmidrule(l){5-6} \cmidrule(l){7-7} \cmidrule(l){8-8} 
\cmidrule(l){9-11} \cmidrule(l){12-13} \cmidrule(l){14-14} \cmidrule(l){15-15} 
 & $Sim \uparrow$ & $CLIP_{i}\uparrow$ & $Shape\downarrow$ & $Expr\downarrow$ & $Pose\downarrow$ & $CLIP_{t}\uparrow$ & $FID\downarrow$ & $Sim \uparrow$ & $CLIP_{i}\uparrow$ & $Shape\downarrow$ & $Expr\downarrow$ & $Pose\downarrow$ & $CLIP_{t}\uparrow$ & $FID\downarrow$ \\ 
\midrule
FlashFace \cite{FlashFace-2024} & 72.1 & 73.3 & 13.8 & 20.6 & 6.7 & 23.4 & 318.7 & 67.5 & \textbf{70.1} & 16.3 & 20.1 & 7.2 & 26.1 & 357.3 \\
IP-Adapter \cite{IP-Adapter-FaceID-2024} & 74.2 & 69.5 & 12.9 & 17.9 & 6.0 & 24.6 & 312.3 & 66.9 & 63.2 & 15.8 & 18.8 & 6.4 & 27.6 & 353.1 \\
PuLID \cite{PuLID-2024} & 73.0 & 69.2 & 13.4 & 19.4 & 6.5 & 25.3 & 301.6 & 69.6 & 64.1 & 14.7 & 19.4 & 6.5 & 27.2 & 337.8 \\
PhotoMaker-V2 \cite{PhotoMaker-V2-2024} & 75.8 & 71.3 & 13.1 & 18.0 & 5.8 & \textbf{26.5} & \textbf{287.4} & 69.0 & 65.0 & 15.2 & 17.2 & 6.0 & 28.0 & 329.5 \\
\textbf{Diff-PC} (Ours) & \textbf{78.3} & \textbf{74.0} & \textbf{10.7} & \textbf{14.5} & \textbf{4.8} & 25.8 & 293.2 & \textbf{72.8} & 69.7 & \textbf{13.1} & \textbf{13.8} & \textbf{5.6} & \textbf{28.3} & \textbf{323.2} \\
\bottomrule
\end{tabular}
\end{adjustbox}
\end{table*}

\paragraph{Evaluation metrics}
We assess the superiority of \method\space from the following perspectives.
\textbf{(1) ID fidelity}: 
evaluated using $Sim$, $CLIP_{i}$ and $Shape$.
In particular, we extract facial embeddings by ArcFace \cite{ArcFace-2019}, where $Sim$ denotes the cosine similarity between the embeddings of reference and generated faces. $CLIP_{i}$ evaluates the CLIP \cite{CLIP-2021} image similarity between the reference face and the generated face. $Shape$ is the root mean square error (RMSE) between the ID parameters of the reference and generated faces. We obtain face parameters using the SMIRK \cite{SMIRK-2024} encoder. 
\textbf{(2) Face controllability}: measured using $Expr$ and $Pose$. Specifically, $Expr$ ($Pose$) represents the RMSE between the target and generated expression (pose) parameters.
\textbf{(3) T2I consistency}: evaluated by $CLIP_{t}$, representing the cosine similarity between the CLIP features of the text prompt and the synthesized image. 
\textbf{(4) Image quality}: 
$FID$ 
\cite{FID-2017} 
(Fréchet Inception Distance) between reference and generated images.
\textbf{(5) Inference time}: time consuming to generate a 1024$\times$1024 image using 8 steps on 1 A100 GPU.

\paragraph{Quantitative comparison}
Tab.~\ref{tab:qualitative} and \ref{tab:ffhq_unsplash} present the quantitative comparison of \method\space with the SOTA methods across three validation sets. All previous methods are configured with Pose-ControlNet \cite{ControlNet-2023} for a fair evaluation. Experimental results demonstrate that our approach outperforms all others in ID fidelity. We attribute this to applying ID-Encoder to extract the refined ID features, as well as the effective injection of ID information via ID-Ctrl and ID-Injector. In addition, our method achieves superior facial controllability, primarily due to the use of 3D-aware facial priors to guide the alignment of ID features during denoising. Lastly, our model also shows an advantage in T2I consistency, thanks to the high-quality dataset we collected.

\begin{figure}[t!]
  \centering
  \includegraphics[width=\linewidth]{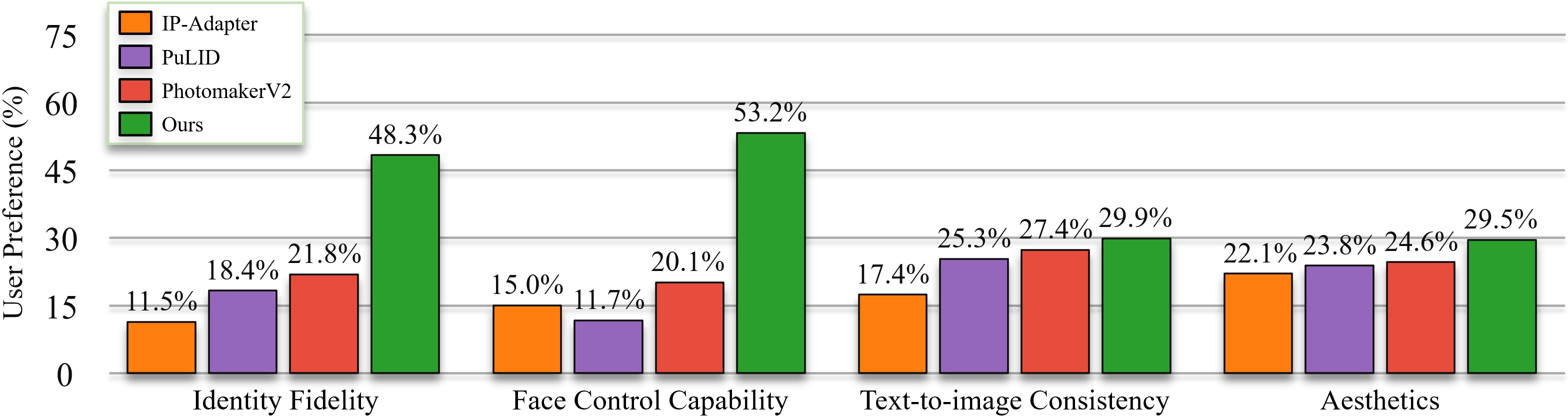}
  \caption{
    \textbf{User study.} Participants consistently favor our method across all four aspects: ID fidelity, face control capability, T2I consistency, and aesthetics.}
  \label{fig:user_study}
\end{figure}

\begin{figure}[t!]
  \centering
  \includegraphics[width=0.85\linewidth]{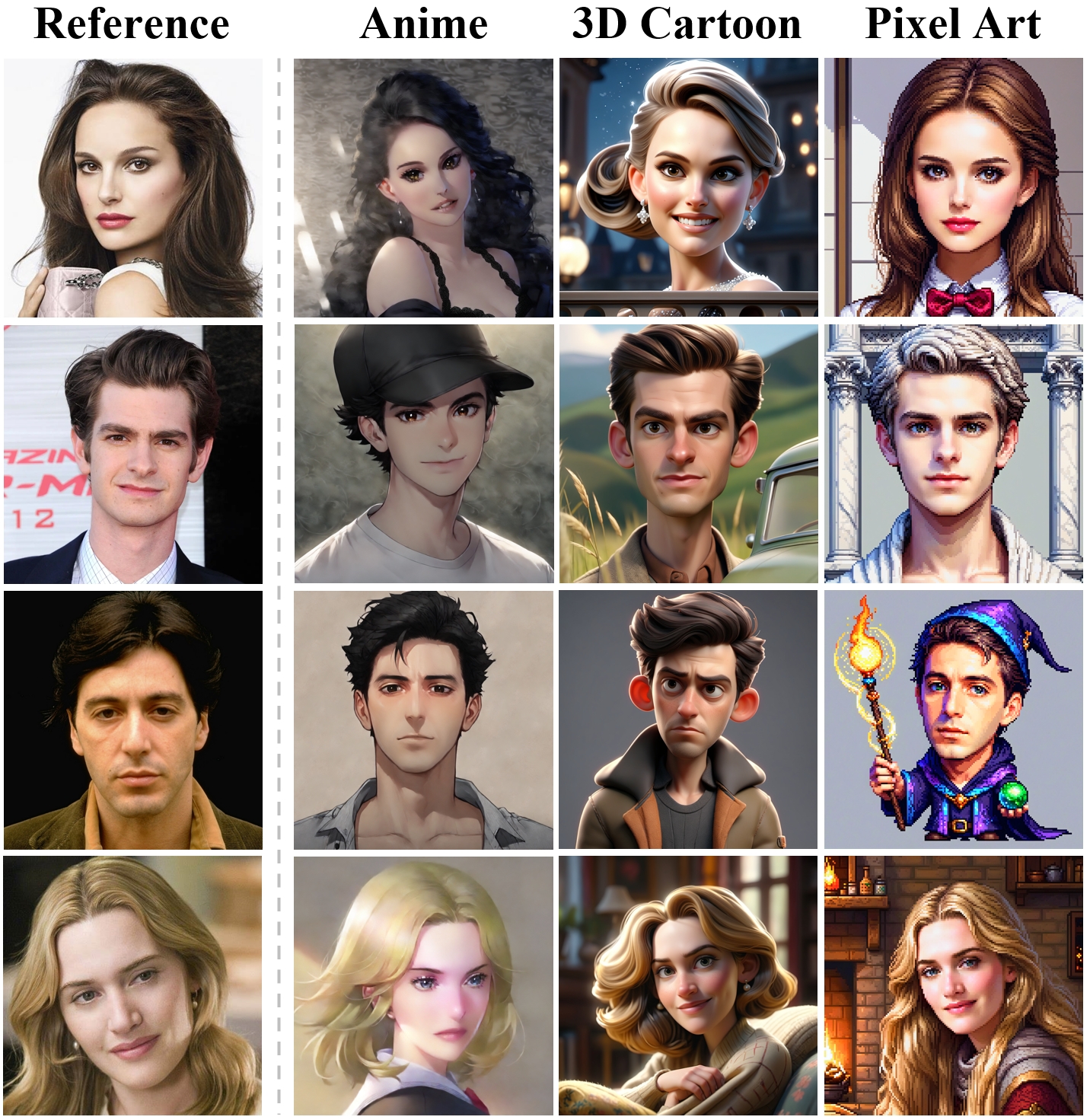}
  \caption{
    \textbf{Compatibility.} \method\space seamlessly adapts to foundation models of varying styles.
  }
  \label{fig:compatibility}
\end{figure}

\paragraph{User study}
To further demonstrate the superiority of \method, we randomly select 30 IDs from ID-centric-200 and generate results using each approach. Then, sixty-three users chose the best images based on the following 4 criteria: ID fidelity, facial controllability, T2I consistency, and aesthetics (image quality). As depicted in Fig.~\ref{fig:user_study}, our model consistently receives the highest preference across all metrics.

\subsection{Further Analysis}

\paragraph{Compatibility} 
Fig.~\ref{fig:compatibility} illustrates multi-style portraits generated by our method, including anime, 3D cartoons, and pixel art. Although \method\space is primarily designed for personalization in real-world scenarios, it can also produce aesthetically pleasing portraits across various styles while effectively preserving the reference ID. This underscores the compatibility of our approach, indicating its seamless adaptability to various foundational models.

\begin{figure}[t!]
  \centering
  \includegraphics[width=\linewidth]{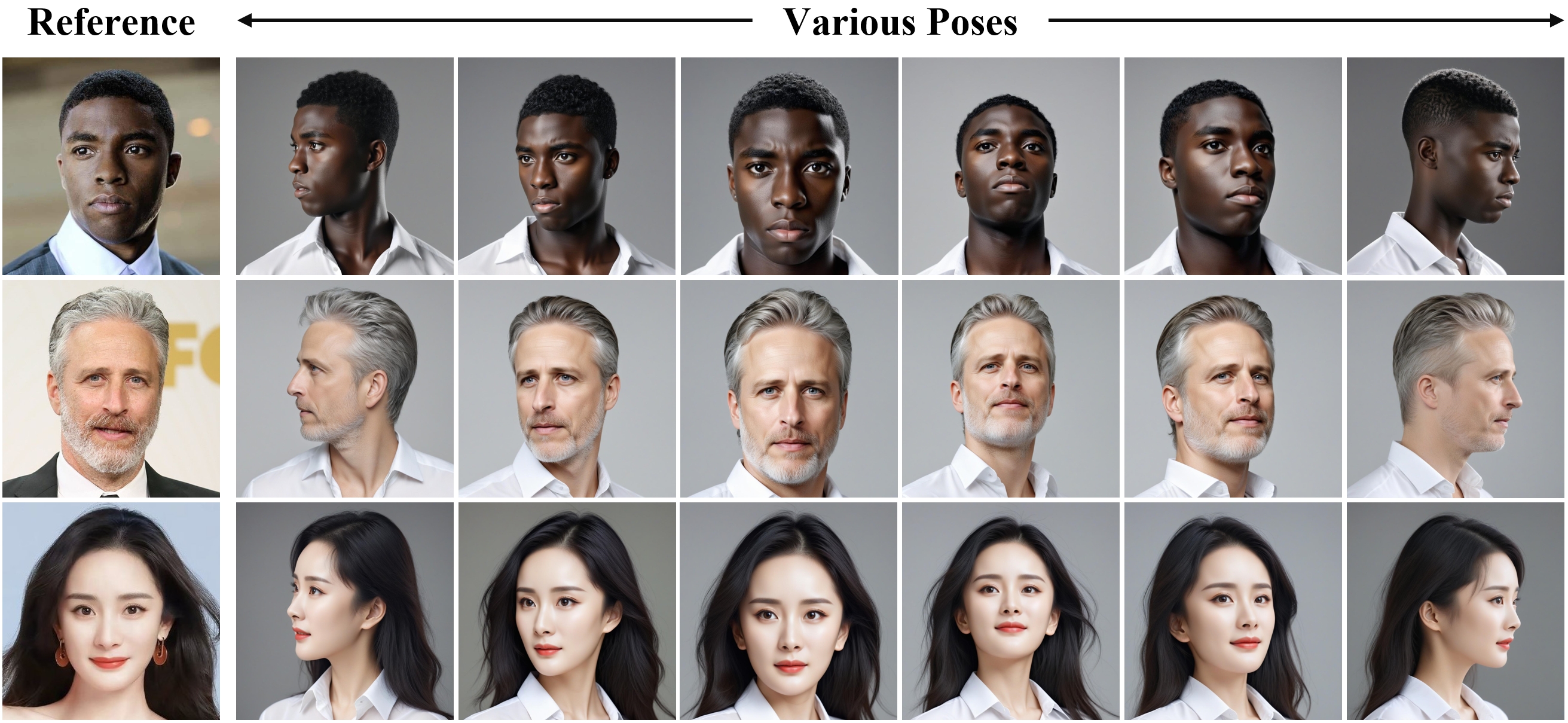}
  \caption{
    \textbf{Pose Controlling} via \method.
  }
  \label{fig:multi_pose}
\end{figure}

\paragraph{Pose Controlling}
Fig.~\ref{fig:multi_pose} presents portraits generated by Diff-PC using 6 different target poses. The results demonstrate that our approach effectively produces images from various viewpoints.

\subsection{Ablation Studies}
To validate the effectiveness of each module, we conduct extensive ablation experiments on ID-centric-200. Following previous studies \cite{PhotoMaker-2023, FlashFace-2024, PuLID-2024}, we reduce the ablation training steps to one-fourth of the original count.

\begin{figure*}[t!]
  \centering
  \includegraphics[width=\linewidth]{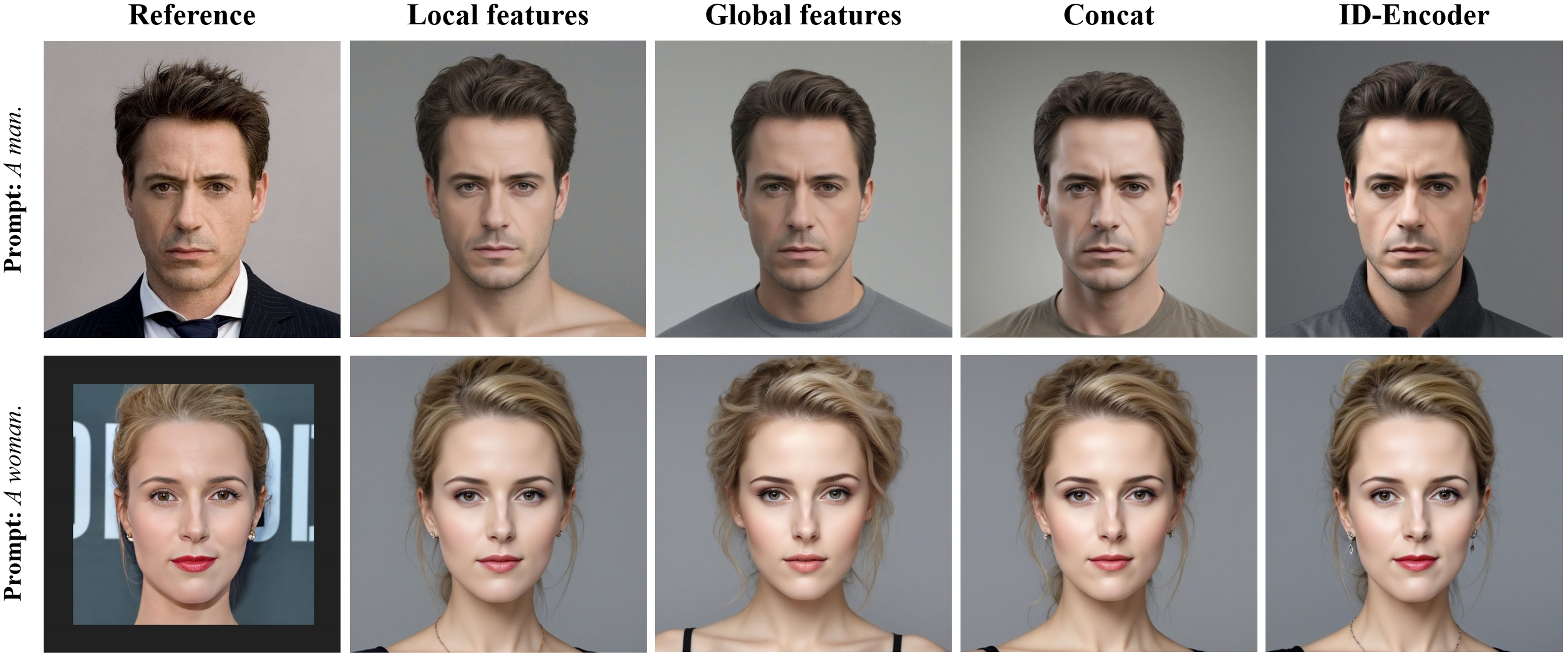}
  \caption{
    \textbf{Effect of ID-Encoder.} \textit{Local} and \textit{global} indicate that local or global facial features are directly fed into ID-Ctrl, respectively. \textit{Concat} refers to concatenating both features before inputting them into ID-Ctrl, without utilizing ID-Encoder.
  }
  \label{fig:vis_id_encoder}
\end{figure*}

\begin{figure*}[t!]
  \centering
  \includegraphics[width=\linewidth]{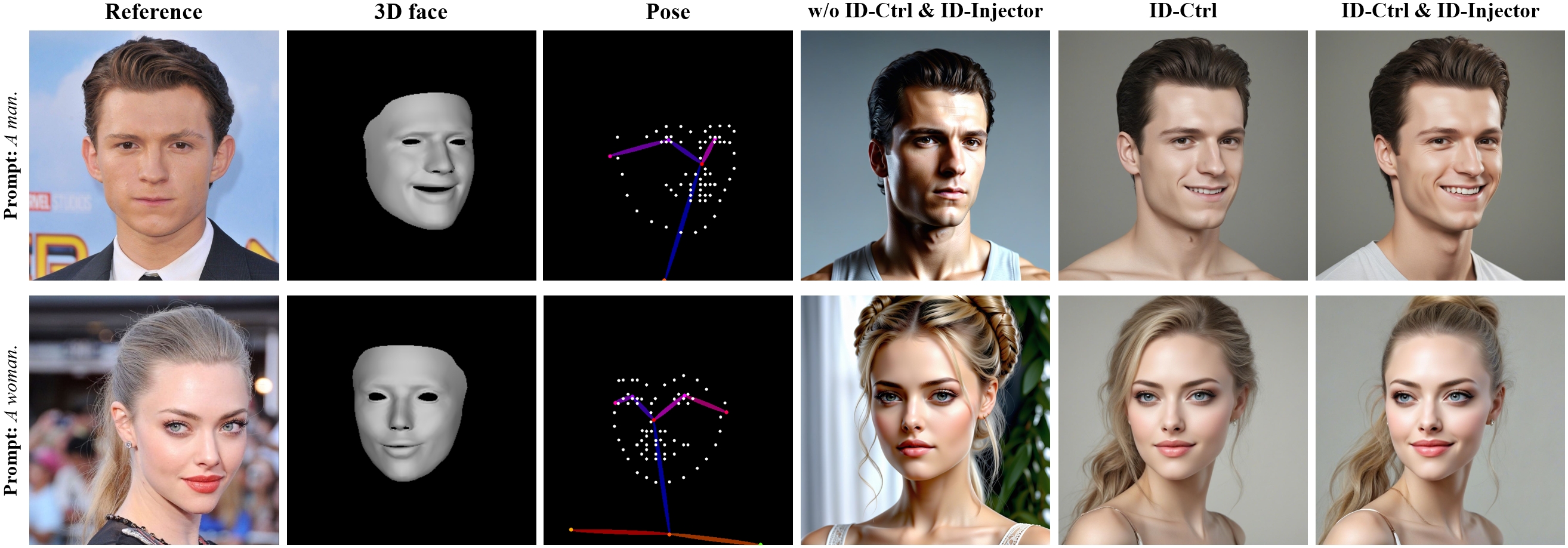}
  \caption{
    \textbf{Impact of ID-Ctrl and ID-Injector.} Column 4 represents using IP-Adapter for ID feature injection and Pose-ControlNet for facial control.
  }
  \label{fig:vis_id_ctrl}
\end{figure*}

\begin{table}[t!]
\caption{
    Ablation on ID-Encoder.
}
\label{tab:id_encoder}
\centering
\begin{adjustbox}{max width=0.8\linewidth}
\begin{tabular}{@{}ccccc@{}}
\toprule
\textbf{Method} & $Sim\uparrow$ & $CLIP_{i}\uparrow$ & $Expr\downarrow$ & $Pose\downarrow$ \\ \midrule
local & 65.2 & 65.5 & 19.8 & 6.6 \\
global & 67.0 & 64.7 & \textbf{19.3} & \textbf{6.3} \\
concat & 67.3 & 65.2 & 19.4 & 6.4 \\
\textbf{ID-Encoder} & \textbf{68.9} & \textbf{66.1} & \textbf{19.3} & 6.4 \\ 
\bottomrule
\end{tabular}
\end{adjustbox}
\end{table}

\begin{table}[t!]
\caption{
    Ablation on ID-Ctrl and ID-Injector.
}
\label{tab:3d_face}
\centering
\begin{adjustbox}{max width=\linewidth}
\begin{tabular}{@{}cccccc@{}}
\toprule
\textbf{ID-Ctrl} & \textbf{ID-Injector} & $Sim\uparrow$ & $CLIP_{i}\uparrow$ & $Expr\downarrow$ & $Pose\downarrow$ \\ \midrule
$\times$ & $\times$ & 62.7 & 62.5 & 21.9 & 7.4 \\
\checkmark & $\times$ & 66.4 & 65.4 & 19.8 & 6.6 \\
\checkmark & \checkmark & \textbf{68.9} & \textbf{66.1} & \textbf{19.3} & \textbf{6.4} \\ \bottomrule
\end{tabular}
\end{adjustbox}
\end{table}

\paragraph{Effect of ID-Encoder}
We investigate the impact of ID-Encoder in Tab.~\ref{tab:id_encoder}. In rows 1 and 2, we employ local or global facial features, respectively. In row 3, we concatenate both features and input them into ID-Ctrl directly. 
Comparing rows 1-3, we observe that local and global features contribute to ID preservation. In addition, combining them can enhance performance, which illustrates their complementary. Comparing rows 3 and 4, utilizing ID-Encoder further retains fine-grained ID details without diminishing face control. In Fig.\ref{fig:vis_id_encoder}, we present visualization results to highlight the effectiveness of ID-Encoder. To facilitate comparison, the reference and target images use the same image.

\begin{table}[t!]
\caption{
    Ablation on ID loss.
}
\label{tab:id_loss}
\centering
\begin{adjustbox}{max width=0.75\linewidth}
\begin{tabular}{@{}ccccc@{}}
\toprule
\textbf{ID loss} & $Sim\uparrow$ & $CLIP_{i}\uparrow$ & $Expr\downarrow$ & $Pose\downarrow$ \\ \midrule
$\times$ & 68.3 & 65.9 & \textbf{19.2} & \textbf{6.4} \\
\checkmark & \textbf{68.9} & \textbf{66.1} & 19.3 & \textbf{6.4}  \\ \bottomrule
\end{tabular}
\end{adjustbox}
\end{table}

\begin{figure*}[t!]
  \centering
  \includegraphics[width=\linewidth]{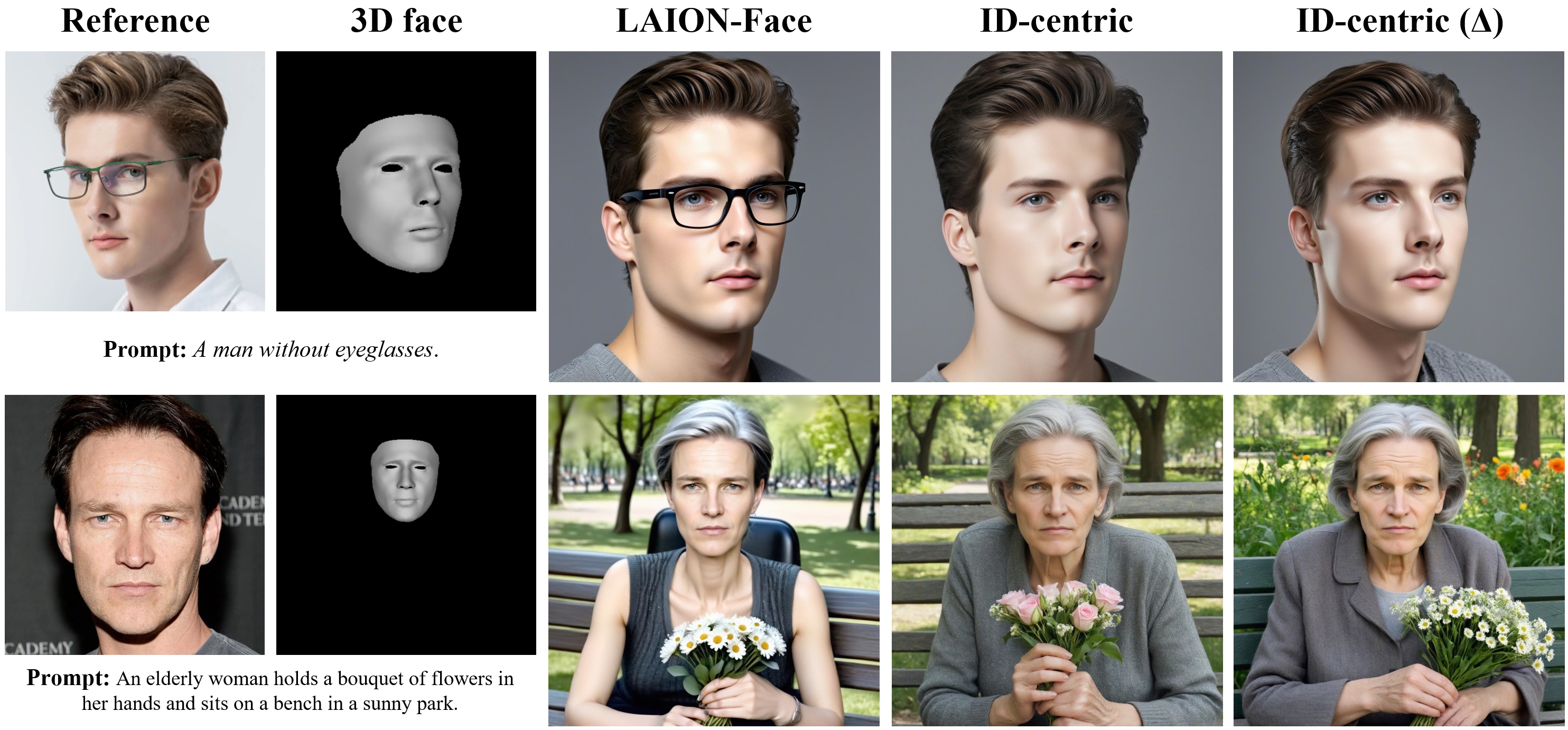}
  \caption{
    \textbf{Impact of different datasets and training strategies.} $\Delta$ denotes using different images from the same ID as reference and target for training.
  }
  \label{fig:vis_t2i}
\end{figure*}

\paragraph{Impact of ID-Ctrl and ID-Injector}
Tab.~\ref{tab:3d_face} demonstrates the impact of ID-Ctrl and ID-Injector. In row 1, we train IP-Adapter to inject ID features, and Pose-ControlNet is used for facial control. Comparing rows 1 and 2, row 1 struggles to control facial attributes using the pose image, and IP-Adapter fails to maintain ID well, leading to subpar results. In contrast, row 2, which integrates ID-Ctrl (with 3D-aware facial priors), significantly improves ID fidelity and facial control precision. Comparing rows 2 and 3, row 3 configures both ID-Ctrl and ID-Injector simultaneously, further enhancing the outcomes. This demonstrates that ID-Injector effectively strengthens the control information provided by ID-Ctrl. Fig.~\ref{fig:vis_id_ctrl} presents comparative images from these experiments, emphasizing the crucial importance of both ID-Ctrl and ID-Injector.

\paragraph{Impact of ID loss}
Tab.~\ref{tab:id_loss} reports impact of using ID loss in our model. Experimental results demonstrate that employing ID loss marginally enhances ID fidelity, while its effect on facial control accuracy is negligible.

\paragraph{ID-centric dataset}
Tab.~\ref{tab:dataset} compares the results of training models with different datasets and strategies. Here, $\Delta$ indicates that the reference and target images use different images from the same ID during training. Row 1 utilizes the unprocessed LAION-Face dataset without $\Delta$. Row 2 employs our collected ID-centric dataset without $\Delta$. Comparing rows 1 and 2,  our dataset filters out blurred and distorted samples, so face similarity and text-to-image alignment improves after training. Comparing rows 2 and 3, the $\Delta$ training strategy further augments face similarity and controllability. This is because the reference and target images are from the same ID in training, whereas they are from different IDs in inference, which causes a particular gap. Row 3 applies $\Delta$ to mitigate this issue and somewhat improves the model's generalization. Fig.~\ref{fig:vis_t2i} presents two comparative examples. The upper example using LAION-Face fails to "\textit{remove eyeglasses}." In the lower case, the ID-centric dataset with $\Delta$ better understands "\textit{an elderly woman}," resulting in a more natural output closely matching the prompt. In both examples, using $\Delta$ achieves the best ID preservation.

\begin{table}[t!]
\caption{
    Ablation on dataset. 
}
\label{tab:dataset}
\centering
\begin{adjustbox}{max width=\linewidth}
\begin{tabular}{@{}ccccccc@{}}
\toprule
\textbf{Dataset} & $\Delta$ & $Sim\uparrow$ & $CLIP_{i}\uparrow$ & $Expr\downarrow$ & $Pose\downarrow$ & $CLIP_{t}\uparrow$ \\ \midrule
LAION-Face & $\times$ & 66.1 & 64.9 & 20.1 & 6.9 & 28.7 \\
ID-centric & $\times$ & 67.8 & 65.7 & 19.8 & 6.8 & \textbf{30.3} \\
\textbf{ID-centric} & \checkmark & \textbf{68.9} & \textbf{66.1} & \textbf{19.3} & \textbf{6.4} & 30.1 \\ \bottomrule
\end{tabular}
\end{adjustbox}
\end{table}

\section{Conclusion}
This paper introduces \method, a diffusion-based pipeline for tuning-free portrait customization. Given one reference image, our method synthesizes realistic ID-preserved portraits, while facial attributes and backgrounds are customized via a target image and text prompt. To achieve efficient ID preservation and face control, we design ID-Ctrl, which leverages the 3D face priors to guide the alignment of ID features. Next, the ID-Injector further reinforces these abilities. In addition, training with our constructed ID-centric dataset also improves ID fidelity and T2I consistency. Comprehensive experiments demonstrate that \method\space outperforms all state-of-the-art models. Moreover, it is compatible with base models of various styles.

\appendix
\section*{Appendix} 
This appendix provides comprehensive dataset statistics and further discussion of \method.

\section{Detailed Dataset Statistics}
\label{app:dataset}
We collect a high-quality, diverse ID-centric dataset for training our proposed \method. This dataset comprises 32k IDs and 650k images, with an average of 20 images per ID. In this section, we report more detailed statistics about this dataset.

\begin{figure}[t!]
  \centering
  \includegraphics[width=\linewidth]{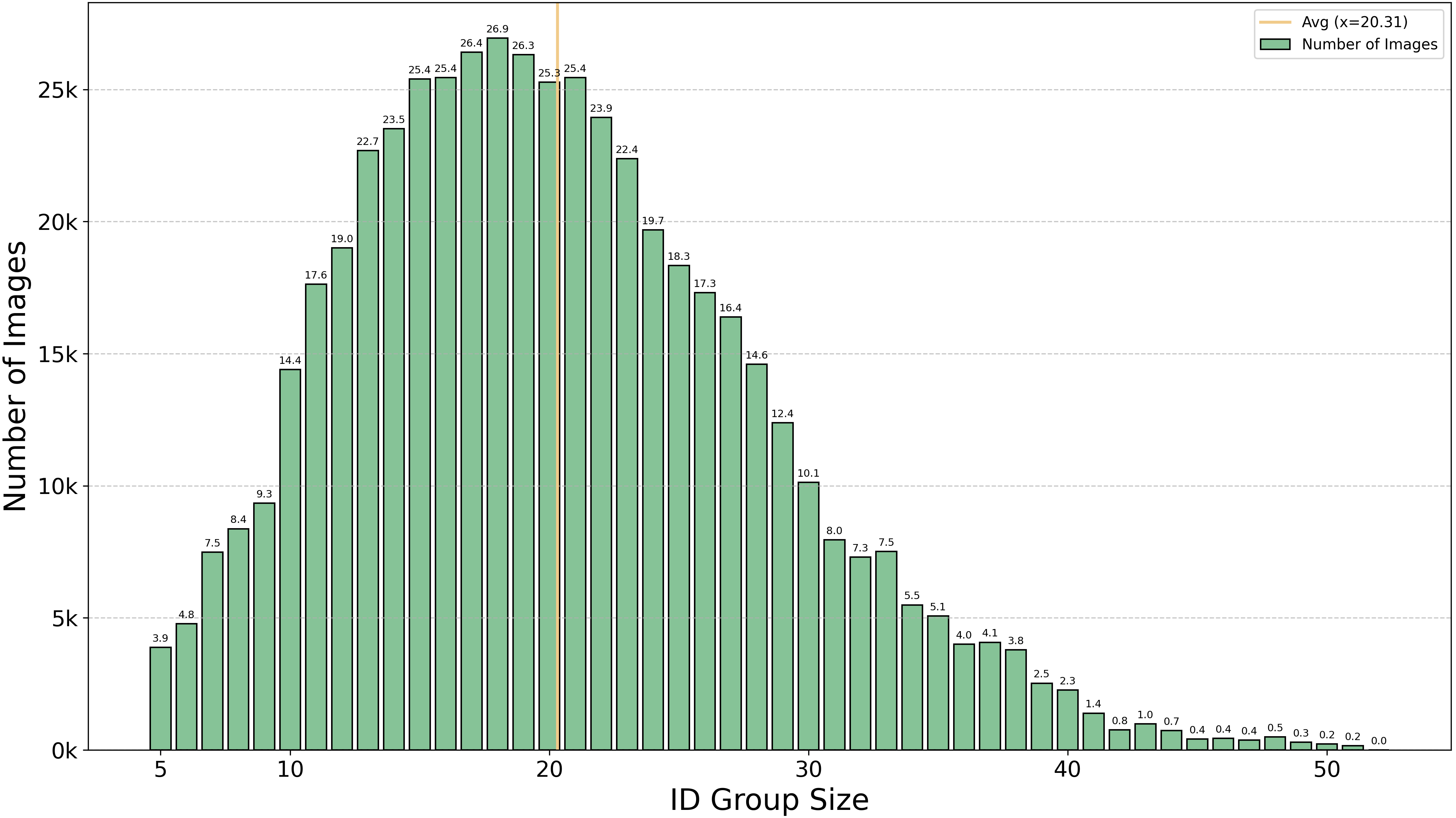}
  \caption{
    \textbf{Distribution of ID group size.}
  }
  \label{fig:id}
\end{figure}

\paragraph{ID group size}
In Fig.~\ref{fig:id}, we present the distribution of image counts in ID groups. ID groups with less than 5 images are filtered out.

\begin{figure}[t!]
  \centering
  \includegraphics[width=\linewidth]{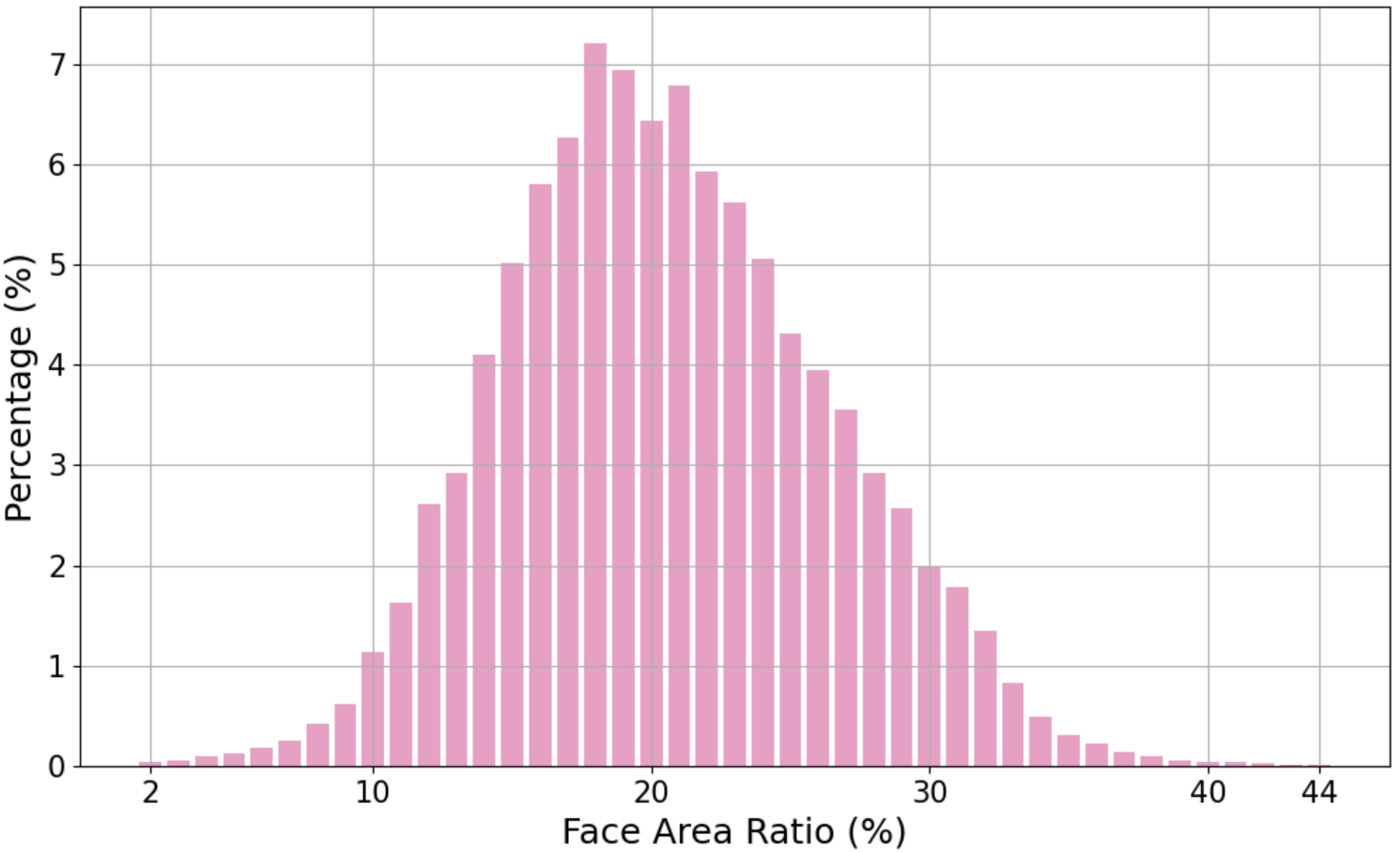}
  \caption{
    \textbf{Statistics of face area ratio.}
  }
  \label{fig:face_area}
\end{figure}

\paragraph{Face area ratio}
Fig.~\ref{fig:face_area} reports the distribution of facial area as a proportion of the entire image, where the facial region is determined using bounding boxes predicted by RetinaFace \cite{RetinaFace-2020}. We observe that images with smaller faces often exhibit blurriness and distortion, leading us to filter out those with facial areas below 2\%. Compared to datasets with face alignment (e.g., FFHQ \cite{FFHQ-2019} and CelebA \cite{dataset-CelebA-2015}), our ID-centric dataset exhibits a broader distribution of facial areas.

\begin{figure}[t!]
  \centering
  \includegraphics[width=\linewidth]{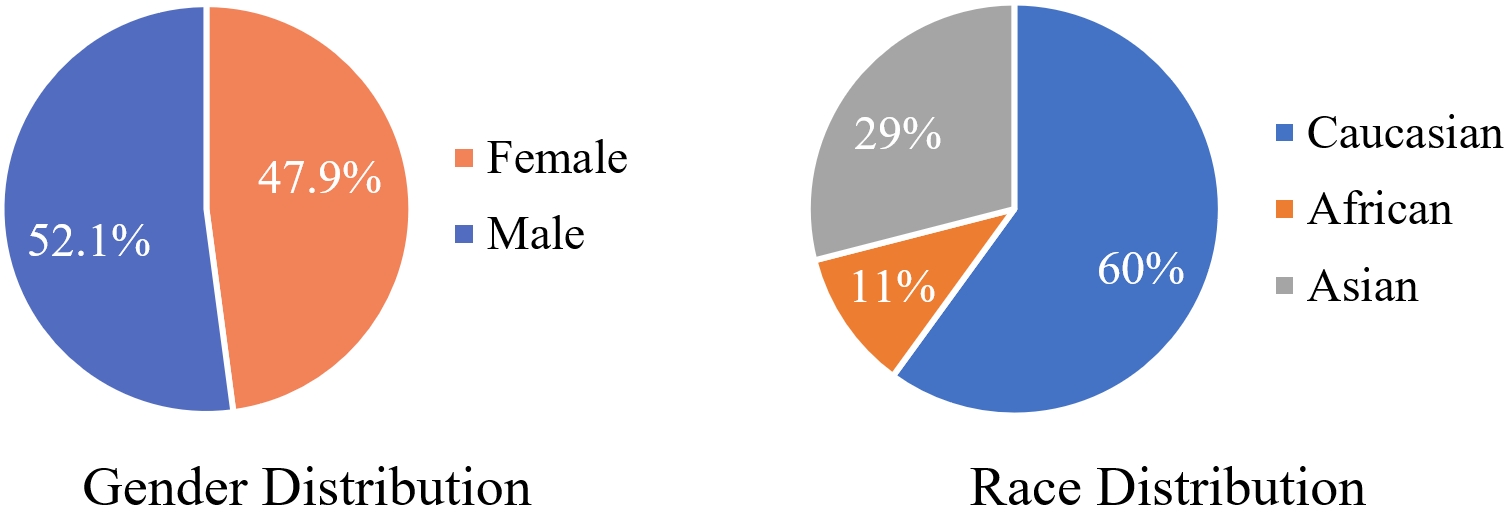}
  \caption{
    \textbf{Gender and race distribution} of IDs in the dataset.
  }
  \label{fig:gender}
\end{figure}

\begin{figure}[t!]
  \centering
  \includegraphics[width=\linewidth]{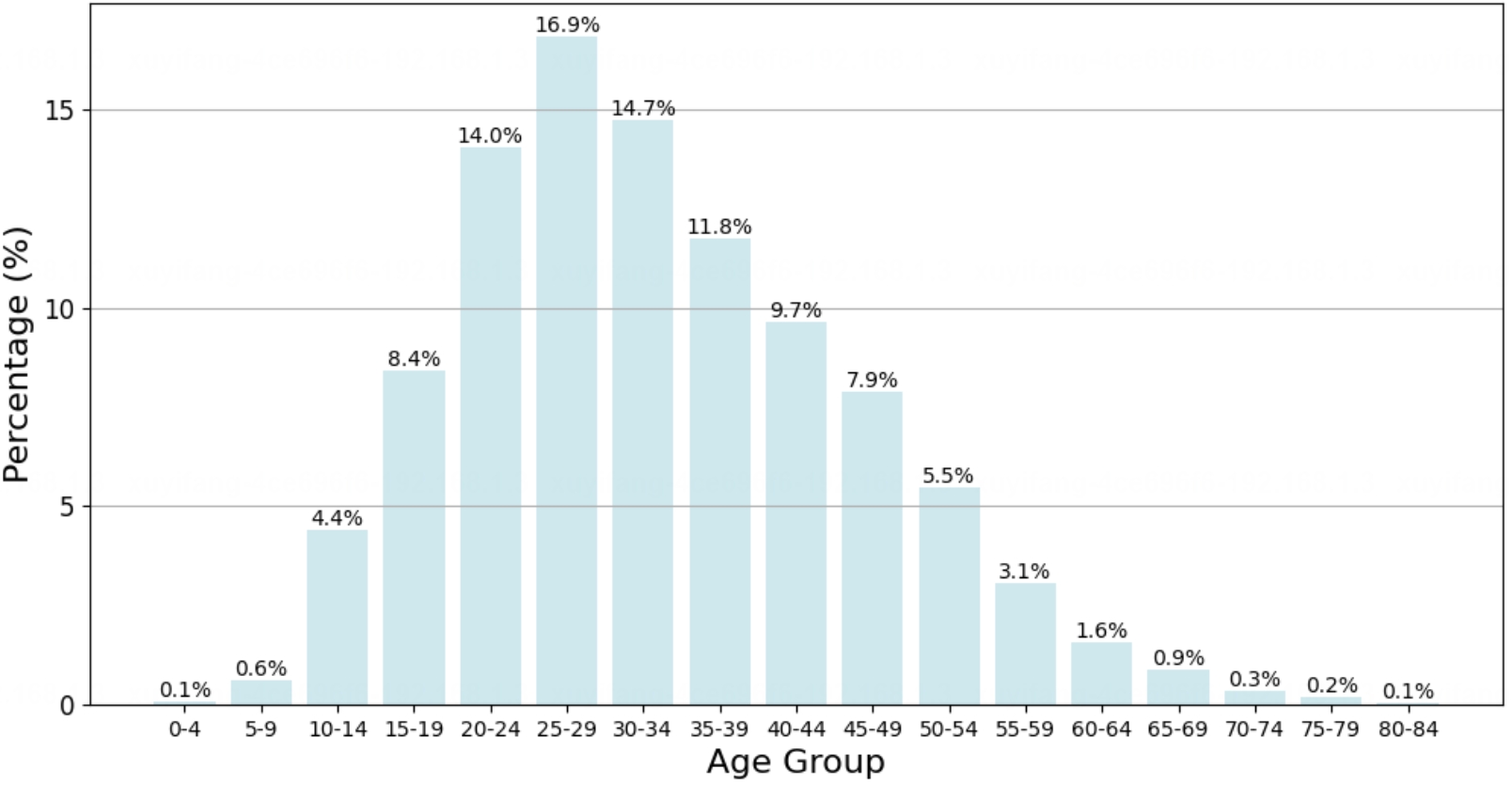}
  \caption{
    \textbf{Age distribution} of IDs.
  }
  \label{fig:age}
\end{figure}

\paragraph{Gender, age and race}
Figs.~\ref{fig:gender}~(left) and \ref{fig:age} show the distributions of gender and age groups, respectively. They are both estimated by Facelib
\footnote{https://github.com/sajjjadayobi/FaceLib}. 
Fig.~\ref{fig:gender}~(right) reports race distribution.

\begin{figure}[t!]
  \centering
  \includegraphics[width=\linewidth]{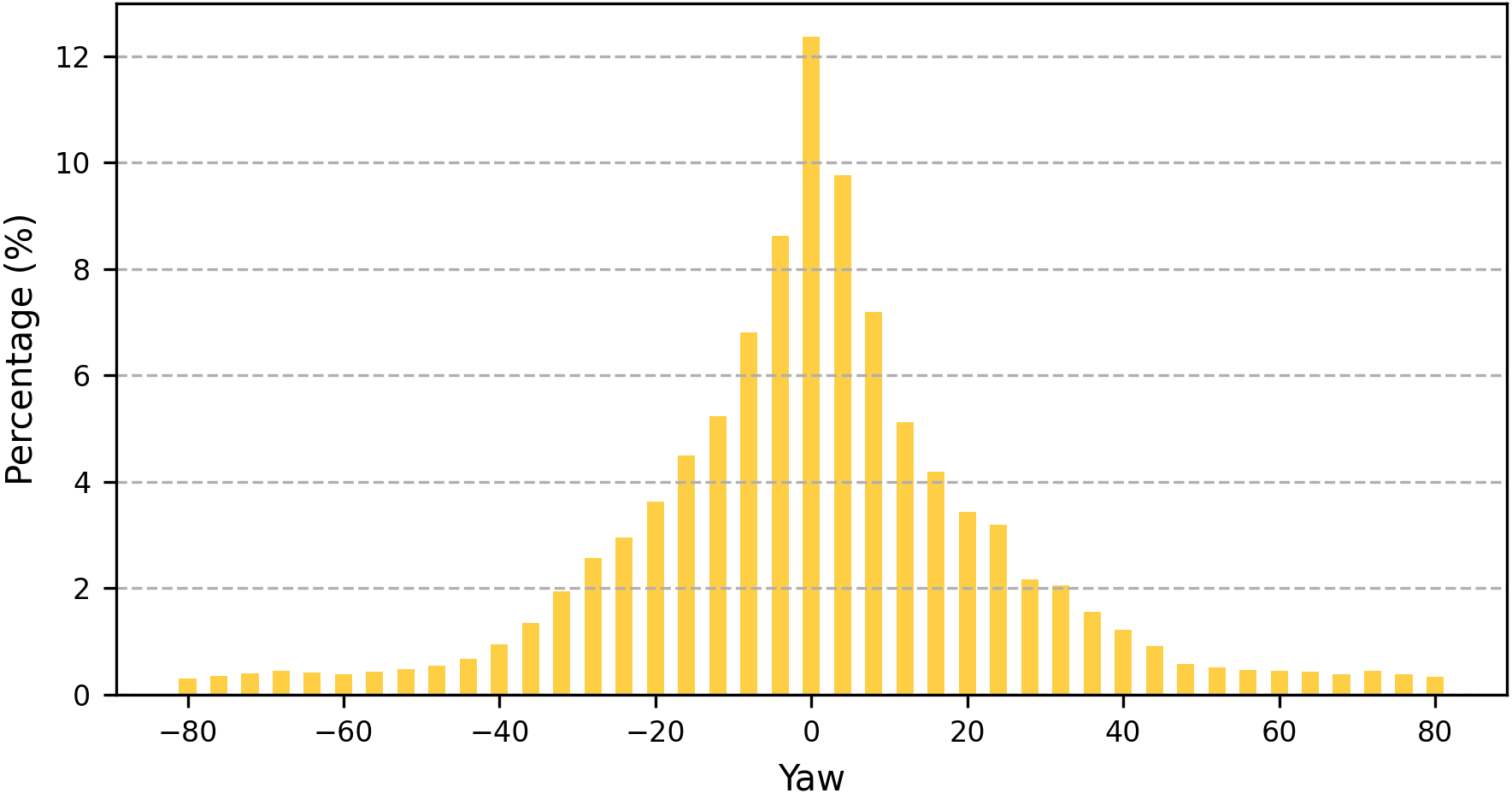}
  \caption{
    \textbf{Yaw distribution} of images.
  }
  \label{fig:yaw}
\end{figure}

\paragraph{Expression and pose}
In Tab.~\ref{tab:expr}, we report the expression distribution of images in the dataset. The expression is identified by ResEmoteNet \cite{ResEmoteNet-2024}. In Fig.~\ref{fig:yaw}, we present the pose (yaw) distribution of images, which estimated using 6DRepNet360 \cite{6DRepNet360-2024}.

\section{Further Discussion}
\label{app:discuss}

\subsection{Limitations}
Although our method has achieved SOTA in qualitative and quantitative comparisons, it still has some limitations. Future research should focus on solving the following problems.

\begin{figure}[t!]
  \centering
  \includegraphics[width=\linewidth]{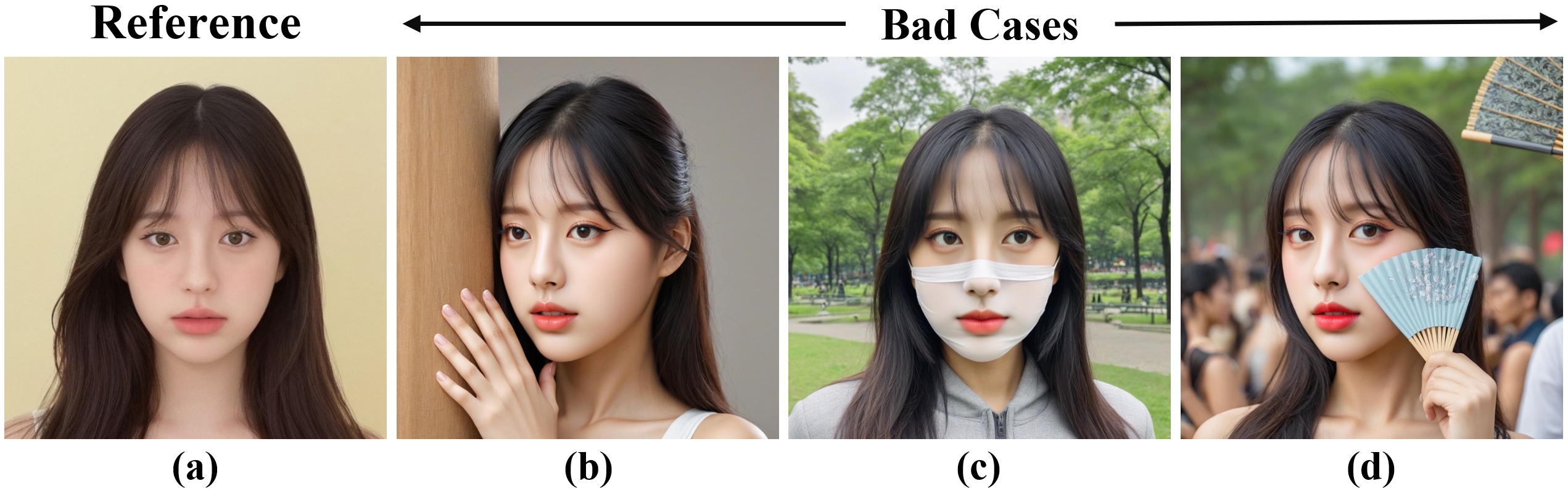}
  \caption{
    \textbf{Some bad cases.} (b) Incorrect hands. (c)(d) Incorrect object occlusion (e.g., mask, fan).
  }
  \label{fig:bad_cases}
\end{figure}

\paragraph{Incorrect hands and occlusion}
As shown in Fig.~\ref{fig:bad_cases}, some errors occur in hand and object occlusion. In Fig.~\ref{fig:bad_cases}~(b), the generated hand has six fingers, because our method models only the face, without considering the hands. In Fig.~\ref{fig:bad_cases}~(c) and (d), incorrect facial occlusions are observed, such as mask and fan. This issue arises because the facial features in Diff-PC are too dominant, leading to T2I consistency.

\begin{figure}[t!]
  \centering
  \includegraphics[width=\linewidth]{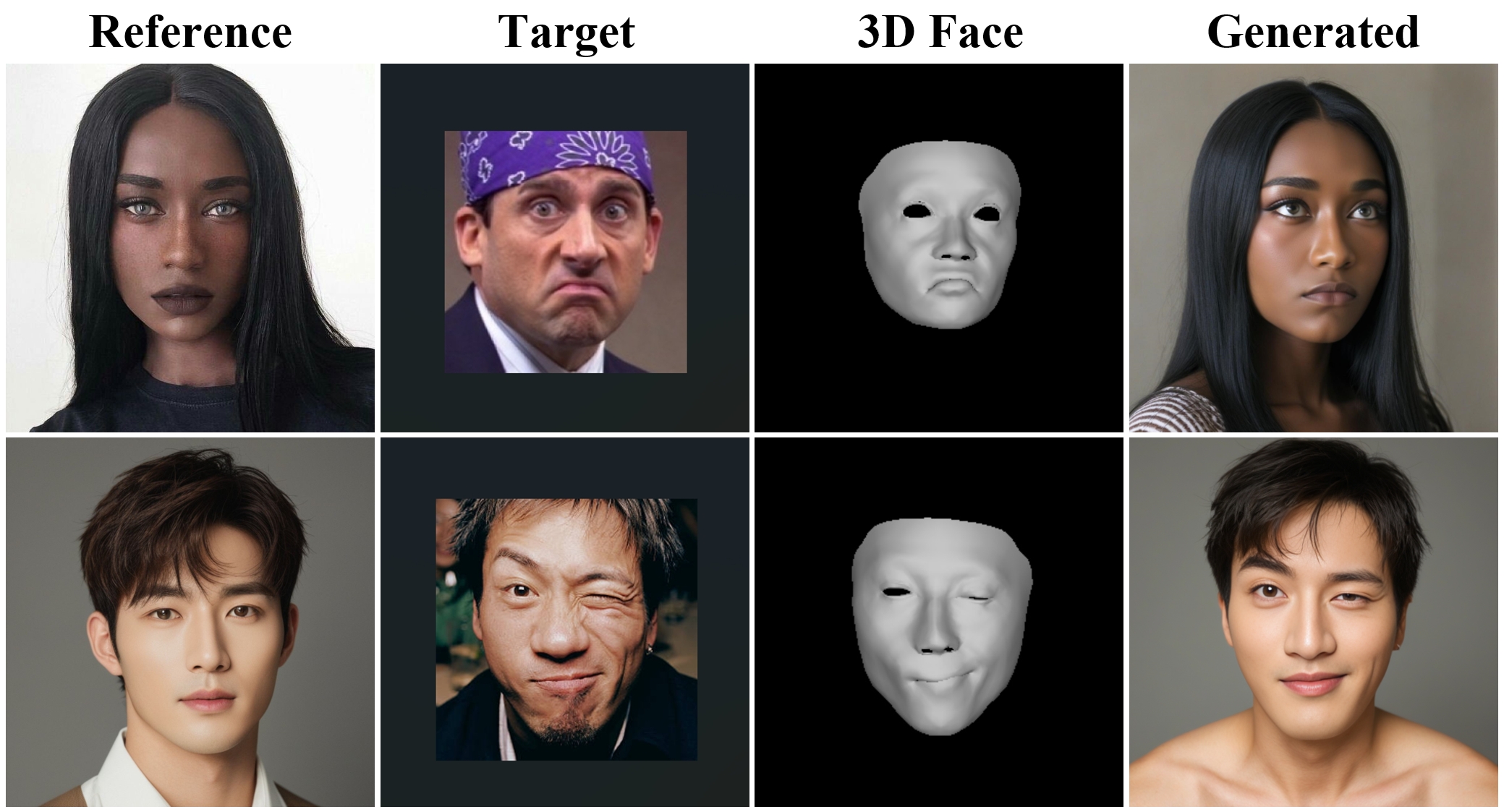}
  \caption{
    Some extreme facial expressions.
  }
  \label{fig:bad_expr}
\end{figure}

\paragraph{Extreme facial expression}
In Fig.~\ref{fig:bad_expr}, we present examples with extreme facial expressions. While the 3D face reconstruction is accurate, the generated expressions still exhibit slight discrepancies with the target image. This is due to the overly strong face preservation in Diff-PC, causing the reference expression to be directly transferred onto the generated image. We plan to address this issue in future work.

\begin{table}
\caption{
    \textbf{Expression distribution} of images in the dataset.
}
\label{tab:expr}
\centering
\begin{adjustbox}{max width=\linewidth}
\begin{tabular}{@{}cccccccc@{}}
\toprule
\textbf{Expression} & Neutral & Happy & Sad & Surprise & Anger & Fear & Disgust \\ 
\midrule
\textbf{Ratio} & 38.6\% & 19.4\% & 13.0\% & 9.8\% & 9.2\% & 6.7\% & 3.3\% 
\\ 
\bottomrule
\end{tabular}
\end{adjustbox}
\end{table}

\subsection{Future works}
The primary experiments in this paper require approximately 1,200 A800 GPU hours, and we lack sufficient computational resources to explore every hyperparameter. Therefore, a promising direction for future research is the acceleration of diffusion model training, such as LCM-LoRA \cite{LCM-LoRA-2023} and SDXL-Lightning \cite{SDXL-Lightning-2024}. Additionally, subsequent studies could explore extending 3D controllable diffusion to a broader range of applications, such as portrait animation \cite{ID-Animator-2024, LivePortrait-2024} and video understanding \cite{han2024video, GPTSee-2023, MH-DETR-2024, VTG-GPT-2023, MRNet-2024, Moment-GPT-2025}.
Finally, experimenting with the latest foundational models like SD3 \cite{SD3-2024} and FLUX-1.1 \cite{Blackforestlabs-FLUX1dev-Hugging-2024} may yield superior results.

\subsection{Broader impacts}
This work contributes a zero-shot portrait customization (PC) framework to the open-source community, enabling image personalization while preserving the reference ID. Depending on the context of the application, this may have positive and negative implications. On the one hand, our proposed \method, based on 3D-aware facial modeling, can propel the development of open-source PC models, facilitating deployment in real-world applications. On the other hand, the generated portrait with high ID fidelity might be misused for face fraud.

\subsection{Ethical considerations}
To achieve a more precise understanding of human anatomy, SDXL \cite{SDXL-2023} incorporates a certain amount of nude human images during training. In our experiments, the generated images may contain nudity if clothing is not specified in the text prompts. Therefore, we recommend enabling NSFW detection to minimize this issue. Aside from this, no other unethical or harmful behavior is observed in the generated images. Finally, it is essential to note that all data and models presented in this paper are intended solely for research purposes, and should not be used for commercial applications.

\bibliographystyle{model1-num-names}

\bibliography{paper}

\end{document}